\documentclass{article}

    \PassOptionsToPackage{numbers, compress}{natbib}
\usepackage[preprint]{neurips_2025}

\usepackage[utf8]{inputenc} % allow utf-8 input
\usepackage[T1]{fontenc}    % use 8-bit T1 fonts
\usepackage{hyperref}       % hyperlinks
\usepackage{url}            % simple URL typesetting
\usepackage{booktabs}       % professional-quality tables
\usepackage{amsfonts}       % blackboard math symbols
\usepackage{nicefrac}       % compact symbols for 1/2, etc.
\usepackage{microtype}      % microtypography
\usepackage{xcolor}         % colors

\usepackage{multirow} 
\usepackage{graphicx}
\usepackage[export]{adjustbox}
\usepackage{float}
\usepackage{epsfig}
\usepackage{graphicx}
\usepackage{amsmath}
\usepackage{amssymb} % http://ctan.org/pkg/amssymb
\usepackage{caption}
\usepackage{xparse} % data cards
\usepackage{wrapfig} % wrapped table
\usepackage{tabularx}
\usepackage{subcaption}
\usepackage{fancyhdr}
\usepackage[hang,flushmargin]{footmisc} % remove indentation in footnote
\usepackage{pifont} % http://ctan.org/pkg/pifont
\usepackage{color}
\usepackage{wrapfig} % wrapped table
\usepackage{boxedminipage}
\usepackage{framed}
\usepackage{soul}
\usepackage{xspace}
\usepackage{booktabs} % professional-quality tables
\usepackage{makecell}
\usepackage{xcolor}
\usepackage{vcell}
\usepackage{colortbl}
\usepackage[nodisplayskipstretch]{setspace}
\usepackage{seqsplit} 
\usepackage{pifont} % circle 1,2,3

\usepackage{listings} % python

\usepackage{algorithmicx, algorithm}
\usepackage[]{algpseudocode}

\definecolor{Red}{RGB}{185, 5, 14}
\definecolor{Green}{RGB}{19,175,85}
\definecolor{Blue}{RGB}{46, 96, 179}

\hypersetup{
    colorlinks=true,
    linkcolor=black,
    citecolor=cyan,
    filecolor=magenta,      
    urlcolor=magenta,
    }

\usepackage{tcolorbox}

\definecolor{myboxcolor}{RGB}{254,254,254} 
\definecolor{myframe}{RGB}{0,0,128} 

\newtcolorbox{mybanner}{
  colback=myboxcolor,
  colframe=myframe,
  boxrule=1pt, % Adjust the border thickness
  left=1pt,
  right=1pt,
  top=1pt,
  bottom=1pt,
}

\newtcolorbox{mybody}{
  colback=myboxcolor,
  colframe=myframe,
  boxrule=1pt, % Adjust the border thickness
  left=1pt,
  right=1pt,
  top=1pt,
  bottom=1pt,
}

\definecolor{my_green}{RGB}{51,102,0}
\definecolor{my_yellow}{RGB}{255,165,0}
\definecolor{my_red}{RGB}{204, 0, 0}

\usepackage{enumitem}

\newcommand{\header}[1]{\text{#1}}

\definecolor{backred}{RGB}{255, 190, 190}
\definecolor{backblue}{RGB}{220, 230, 250}

\definecolor{lightblue}{RGB}{225, 233, 249} 
\definecolor{lightred}{RGB}{254, 219, 219} 
\usepackage{tikz}

\definecolor{shadecolor}{RGB}{230,230,230}

\newcommand{\cello}[0]{\textbf{\texttt{Cell-o1}}}
\newcommand{\bench}[0]{\textbf{\texttt{CellPuzzles}}}

\newcommand{\beginthink}[0]{\textbf{\texttt{\textcolor{Blue}{<}\textcolor{Blue}{t}\textcolor{Blue}{h}\textcolor{Blue}{i}\textcolor{Blue}{n}\textcolor{Blue}{k}\textcolor{Blue}{>}}}}

\newcommand{\think}[0]{\textbf{\texttt{\textcolor{Blue}{<}\textcolor{Blue}{/}\textcolor{Blue}{t}\textcolor{Blue}{h}\textcolor{Blue}{i}\textcolor{Blue}{n}\textcolor{Blue}{k}\textcolor{Blue}{>}}}}

\newcommand{\beginanswer}[0]{\textbf{\texttt{\textcolor{Red}{<}\textcolor{Red}{a}\textcolor{Red}{n}\textcolor{Red}{s}\textcolor{Red}{w}\textcolor{Red}{e}\textcolor{Red}{r}\textcolor{Red}{>}}}}

\newcommand{\answer}[0]{\textbf{\texttt{\textcolor{Red}{<}\textcolor{Red}{/}\textcolor{Red}{a}\textcolor{Red}{n}\textcolor{Red}{s}\textcolor{Red}{w}\textcolor{Red}{e}\textcolor{Red}{r}\textcolor{Red}{>}}}}

\newcommand{\thinkbegin}{\beginthink\xspace}
\newcommand{\thinkend}{\think\xspace}
\newcommand{\answerbegin}{\beginanswer\xspace}
\newcommand{\answerend}{\answer\xspace}
\newcommand{\Ours}{\cello\xspace}
\newcommand{\Bench}{\bench\xspace}
\usepackage{enumitem}
\usepackage{graphicx}

\title{
\Ours: Training LLMs to Solve Single-Cell Reasoning Puzzles with Reinforcement Learning}

\author{
    \textbf{Yin Fang}$^1$\footnotemark[1] \quad  
    \textbf{Qiao Jin}$^1$\thanks{$\quad$ Equal Contribution.} \quad
    \textbf{Guangzhi Xiong}$^2$ \quad 
    \textbf{Bowen Jin}$^3$  \quad 
    \textbf{Xianrui Zhong}$^3$ \quad \\ 
    \textbf{Siru Ouyang}$^{3}$ \quad 
    \textbf{Aidong Zhang}$^{2}$ \quad 
    \textbf{Jiawei Han}$^{3}$ \quad 
    \textbf{Zhiyong Lu}$^{1}$\thanks{$\quad$ Corresponding Author.}
  \\ 
  $^1$ National Institutes of Health \quad
  $^2$ University of Virginia \\
  $^3$ University of Illinois Urbana-Champaign  
}

\begin{document}

\vspace{-0.3cm}

\maketitle

\vspace{-0.5cm}

\begin{abstract}
\vspace{-0.2cm}

Cell type annotation is a key task in analyzing the heterogeneity of single-cell RNA sequencing data.
Although recent foundation models automate this process, they typically annotate cells independently, without considering batch-level cellular context or providing explanatory reasoning.
In contrast, human experts often annotate distinct cell types for different cell clusters based on their domain knowledge.
To mimic this workflow, we introduce the \Bench task, where the objective is to assign unique cell types to a batch of cells.
This benchmark spans diverse tissues, diseases, and donor conditions, and requires reasoning across the batch-level cellular context to ensure label uniqueness.
We find that off-the-shelf large language models (LLMs) struggle on \Bench, with the best baseline (OpenAI's \texttt{o1}) achieving only 19.0\% batch-level accuracy.
To fill this gap, we propose \Ours, a 7B LLM trained via supervised fine-tuning on distilled reasoning traces, followed by reinforcement learning with batch-level rewards. 
\Ours achieves state-of-the-art performance, outperforming \texttt{o1} by over 73\% and generalizing well across contexts.
Further analysis of training dynamics and reasoning behaviors provides insights into batch-level annotation performance and emergent expert-like reasoning. 
Code and data are available at \url{https://github.com/ncbi-nlp/cell-o1}.
\end{abstract}

\vspace{-0.3cm}
\begin{figure}[h!]
\centering
\includegraphics[width=0.96\textwidth]{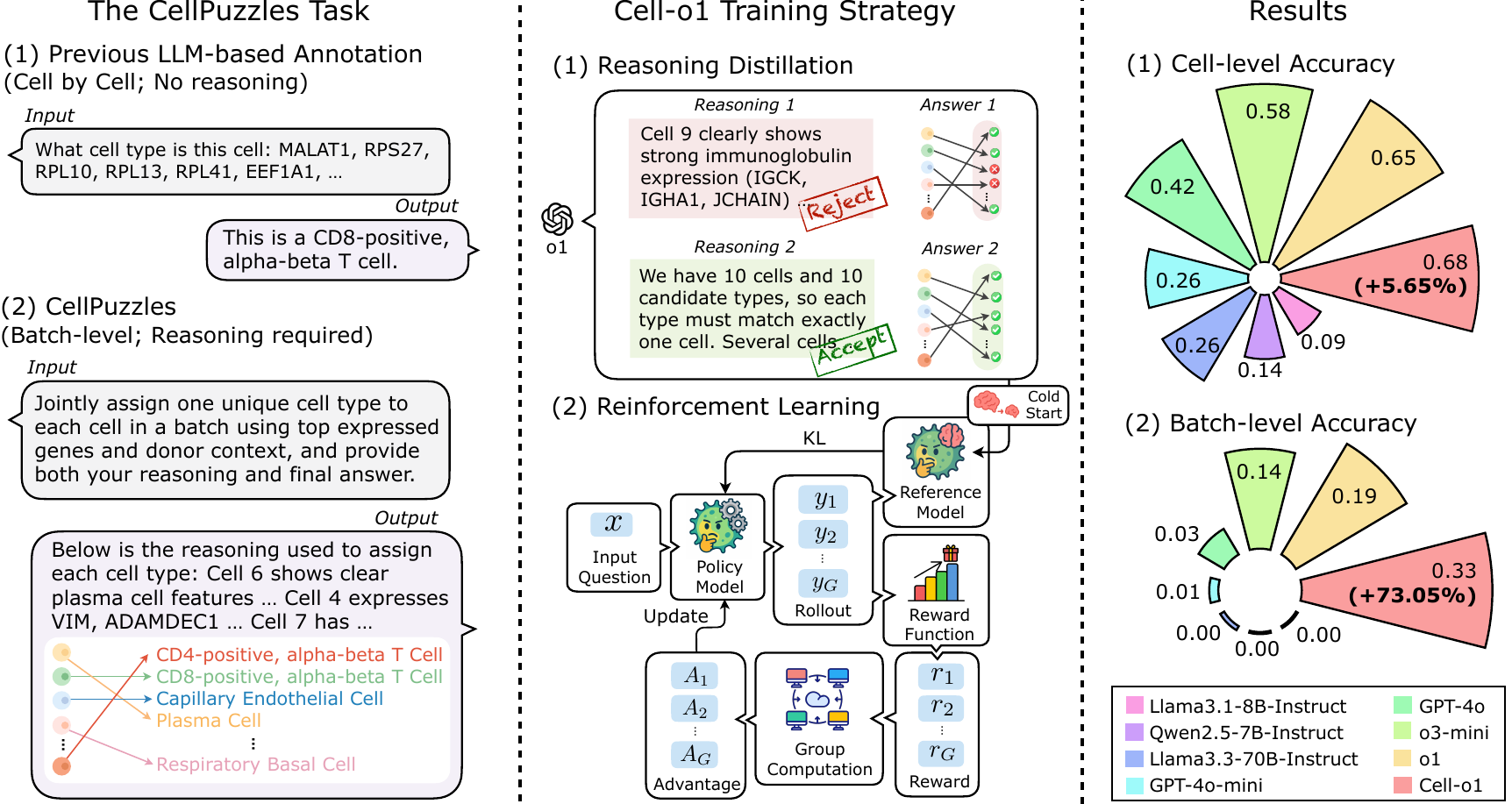}
\caption{Overview of this work. \Ours achieves state-of-the-art on the \Bench task.}
\label{fig:overview}
\end{figure}
\vspace{-0.5cm}

\section{Introduction}
\label{sec:intro}
Assigning accurate cell types to single-cell RNA sequencing (scRNA-seq) profiles is fundamental for understanding biological heterogeneity across tissues, diseases, and individuals~\cite{kiselev2019challenges,luecken2019current}.
Traditional annotation workflows rely heavily on expert knowledge, typically involving clustering to group similar cells~\cite{butler2018integrating,traag2019louvain}, followed by manual inspection of marker gene expression to assign cell type labels based on biological domain knowledge~\cite{stuart2019comprehensive,aran2019reference}.
While accurate, this process is time-consuming and labor-intensive, with limited scalability across large or new datasets~\cite{kiselev2019challenges,luecken2019current,stuart2019comprehensive,lahnemann2020eleven,abdelaal2019comparison}.

Recent advances in deep learning have significantly improved automated cell type annotation through the development of single-cell foundation models~\cite{yang2022scbert, theodoris2023transfer, hao2024large}. These models leverage large-scale unsupervised pre-training to capture complex gene-expression patterns, enabling better representation learning and enhanced performance across various downstream tasks. In parallel, large language models (LLMs) have also been adapted for single-cell applications, either by translating gene-expression profiles into textual representations~\cite{hou2024assessing,levine2023cell2sentence,liu2023scelmo,chen2024simple} or by integrating multimodal gene embeddings~\cite{zhao2024langcell,schaefer2024multimodal,fang2025multi, shi2025multimodal}. However, both foundation models and LLMs typically annotate each cell independently without considering the shared biological context or batch-level gene-expression information, which fundamentally differs from expert annotation practices~\cite{pasquini2021automated}. Furthermore, most automated methods directly predict cell types without articulating the underlying reasoning process, making their decisions difficult to interpret and validate~\cite{azodi2020opening}.

To bridge this gap, we first introduce \Bench, a novel benchmark that formulates cell type annotation as a batch-level reasoning task, closely mimicking expert annotation workflows. 
As illustrated in Figure~\ref{fig:overview}, unlike traditional methods that label each cell independently, \Bench requires jointly assigning unique labels to all cells in a batch, using both their gene-expression profiles and shared contextual metadata. Extensive evaluations show that state-of-the-art LLMs struggle with this formulation, with the best-performing model (OpenAI's \texttt{o1}~\cite{jaech2024openai}) achieving only 19.0\% batch-level accuracy, reflecting the complexity of this task.

To address this challenge, we propose \Ours, a reasoning-enhanced LLM trained in two phases as shown in Figure~\ref{fig:overview}: supervised fine-tuning (SFT) on expert-like reasoning traces distilled from a frontier LLM to guide structured and interpretable decision-making; followed by reinforcement learning (RL) with batch-level rewards to encourage consistent, context-aware label assignments.
On \Bench, \Ours outperforms all baseline models in both cell-level and batch-level accuracy. 
Further analyses of training dynamics, reasoning behavior, and prediction errors offer insights into the model’s generalization, interpretability, and reasoning ability.
Notably, \Ours exhibits emergent behaviors such as self-reflection~\cite{renze2024self,zhaoevaluating}, where the model revisits and revises earlier predictions, and curriculum reasoning~\cite{cornacchia2023mathematical,maharana2022curriculum}, where it prioritizes simpler cases before tackling harder ones—both resembling strategies used by human experts.
In summary, our key contributions include:
\begin{itemize}[leftmargin=*]
    \item We introduce \Bench, a novel benchmark that reformulates cell type annotation as a batch-level reasoning task, requiring the joint assignment of unique cell types to each cell based on both expression profiles and contextual metadata.
    \item We propose \Ours, a reasoning-enhanced LLM trained via SFT on distilled reasoning traces and RL with batch-level rewards, enabling interpretable annotation across cell batches.
    \item We conduct extensive analyses of training dynamics, model behaviors, and error patterns, revealing emergent expert-like reasoning capabilities and robust performance on unseen biological conditions.
\end{itemize}

\section{Related Work}

\textbf{Traditional Cell Type Annotation.}
Manual cell type annotation typically follows a multi-step process combining clustering, marker gene identification, and reference-based assignment. Cells are first grouped using unsupervised clustering algorithms such as hierarchical clustering, Louvain, or Leiden~\cite{butler2018integrating,traag2019louvain}. Experts then inspect the resulting clusters, identify distinguishing marker genes, and assign cell type labels based on prior biological knowledge and curated reference databases~\cite{stuart2019comprehensive,aran2019reference}. The annotation process often involves iterative refinement, requiring the integration of various data sources and repeated evaluations.
Although effective, expert-driven annotation is slow, labor-intensive, and requires substantial domain expertise. It also suffers from limited reproducibility and scalability when applied to large or multi-batch datasets~\cite{stuart2019comprehensive,abdelaal2019comparison}. Moreover, its inability to systematically incorporate biological context further limits robustness and interpretability~\cite{trapnell2015defining}.

\textbf{Single-Cell Foundation Models.}
Recent advancements in deep learning have led to the emergence of powerful single-cell foundation models designed to capture complex gene expression patterns through deep representation learning~\cite{yang2022scbert, theodoris2023transfer, hao2024large}. 
Built upon transformer architectures, these models leverage extensive unsupervised pre-training on large-scale single-cell transcriptomic datasets to achieve enhanced representation quality and downstream task performance. These foundation models have demonstrated remarkable capabilities in transferring learned knowledge across different datasets and tasks, including cell type annotation~\cite{yang2022scbert,theodoris2023transfer}, gene network inference~\cite{van2020scalable,osorio2020sctenifoldnet}, and perturbation prediction~\cite{lotfollahi2019scgen,ji2021machine}. 
Nevertheless, these models typically operate at the single-cell level, independently annotating each cell. Their performance in zero-shot settings remains limited, often necessitating fine-tuning with additional labeled data~\cite{yang2022scbert,cui2024scgpt,theodoris2023transfer}.

\textbf{LLM-driven Single-Cell Annotation.}
Recent research efforts have explored adapting single-cell gene expression data for LLMs, thereby leveraging their strong instruction-following, representation learning, and generalization capabilities~\cite{achiam2023gpt,touvron2023llama}. With their instruction-following and in-context learning capabilities, LLMs can quickly adapt to different single-cell tasks, an advantage that traditional single-cell analysis methods cannot achieve~\cite{lou2024large,brown2020language}. 
To bridge the modality gap between scRNA-seq data and language models, several strategies have been proposed, including ranking genes by expression to form ordered lists~\cite{hou2024assessing,levine2023cell2sentence}, embedding genes based on curated textual descriptions~\cite{liu2023scelmo,chen2024simple}, and directly encoding raw expression profiles using multimodal architectures~\cite{zhao2024langcell,schaefer2024multimodal,fang2025multi, shi2025multimodal}. 
These approaches enable LLMs to process biological information and perform tasks such as cell type annotation within a textual or multimodal framework.
While promising, existing LLM-driven approaches still face challenges. Most approaches process cells individually, not fully exploiting batch-level context information. Moreover, they output annotation results without explicitly articulating the underlying reasoning process~\cite{wei2022chain,kojima2022large}, limiting interpretability and biological trustworthiness. 

\section{\Bench: Batch-Level Reasoning for Cell Type Annotation}
\label{sec:bench}

In real-world single-cell analysis, cell type annotation is rarely performed on individual cells in isolation. Instead, it typically occurs at the \textit{batch level}, where groups of cells from the same donor or sample are jointly analyzed.  Experts commonly cluster cells based on gene expression profiles, identify representative marker genes for each cluster, and assign labels by integrating both expression patterns and contextual metadata such as tissue origin and disease status.

\begin{figure}[h!]
\centering
\includegraphics[width=\textwidth]{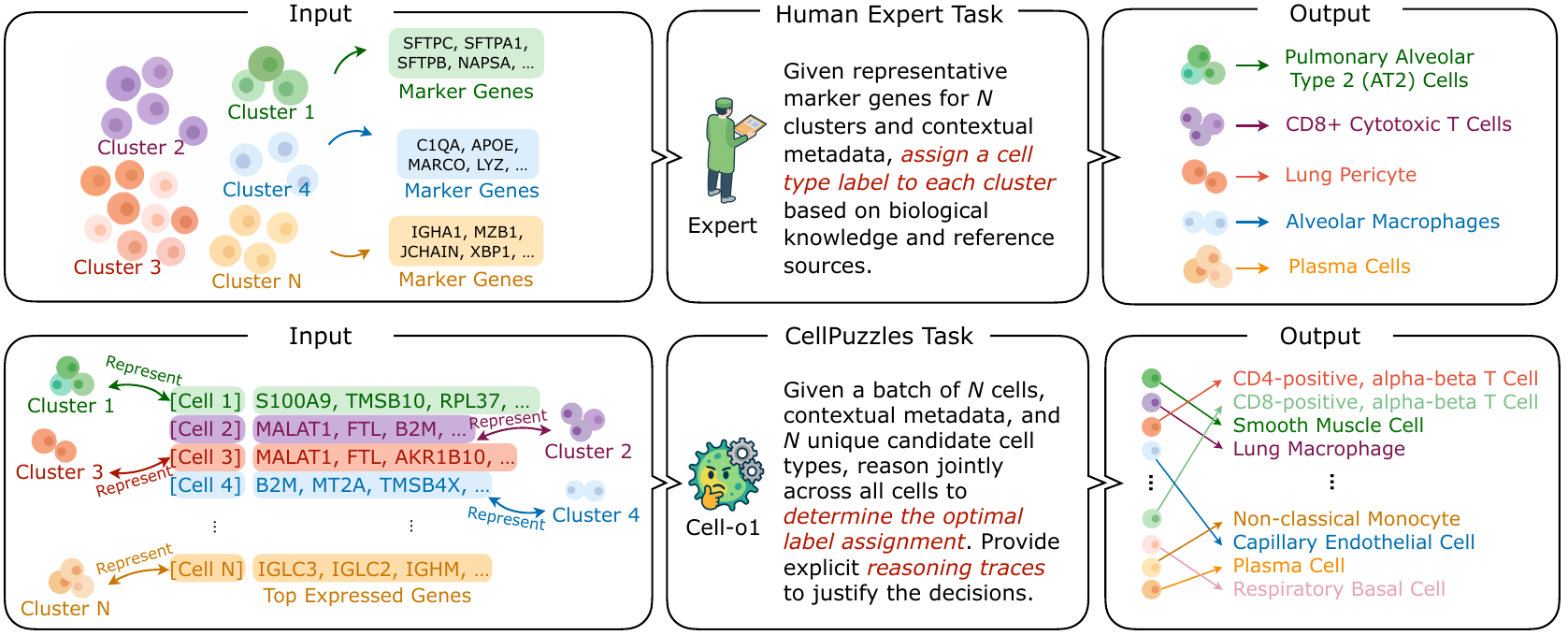}
\caption{
\Bench formulates cell type annotation as a batch-level reasoning task that integrates gene expression and contextual metadata, inspired by how experts annotate cells in practice.
}
\label{fig:cellpuzzles}
\end{figure}

To emulate this expert-driven process, we introduce \Bench, a novel benchmark shown in Figure~\ref{fig:cellpuzzles} that formulates cell type annotation as a batch-level reasoning task. 
Formally, each instance consists of a batch of $N$ cells $\mathcal{C} = \{c_1, c_2, \dots, c_N\}$, each sampled from a distinct cell type within the same donor and experimental batch.
Each cell $c_i$ serves as a proxy for a cluster centroid and is represented by a ranked list of its top $M$ expressed genes $g_i = [g_{i1}, g_{i2}, \dots, g_{iM}]$, which approximate the differentially expressed genes used by experts to define and annotate single cell clusters.

The entire batch is associated with a contextual description $m$, derived from donor-level metadata such as tissue type, disease status, sex, developmental stage, and other biologically relevant attributes when available. % (e.g., smoking status or lung condition). 
We also provide a candidate label set $\mathcal{Y} = \{y_1, y_2, \dots, y_N\}$, which contains the ground-truth cell types for the $N$ cells in the batch, randomly shuffled to remove positional biases. This setup mirrors the expert annotation scenario, where cell types are selected from a biologically plausible candidate pool rather than freely generated.
The objective is to learn a mapping $f: \mathcal{C} \rightarrow \mathcal{Y}$ that assigns each cell to a unique label in $\mathcal{Y}$ by jointly considering gene expression profiles and contextual metadata, while generating interpretable reasoning traces to justify the label assignments.

\begin{figure}[h!]
\centering
\includegraphics[width=1.0\textwidth]{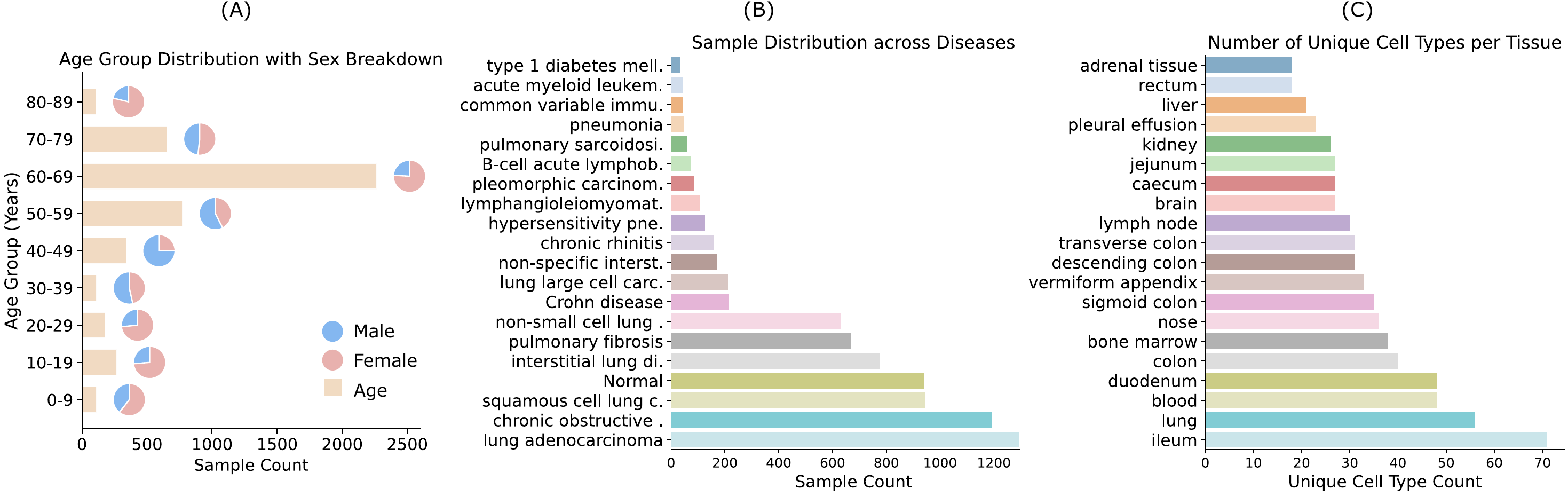}
\caption{
Data distribution of \Bench.
(A) Age group distribution with male-female breakdown.
(B) Sample distribution across a wide range of disease conditions.
(C) Number of unique cell types observed across different tissues. 
For clarity, only the top 20 entries are shown in (B) and (C).
}
\label{fig:data_distribution}
\end{figure}

As illustrated in Figure~\ref{fig:data_distribution}, \Bench is built from human single-cell datasets in Cell$\times$Gene portal~\cite{czi2025cz} that span diverse tissues, disease conditions, and donor demographics. 
The detailed dataset construction protocol is provided in the Supplementary Materials.

\section{\Ours: Large Reasoning Model for Cell Batch Annotation}

Built upon \Bench, we introduce \Ours, a large reasoning model that emulates expert annotation strategies for batch-level single-cell analysis. It performs reasoning over top-expressed genes and shared biological context to assign unique cell type labels from a given candidate set.

Figure~\ref{fig:overview} illustrates our training strategy.
We begin by introducing a unified prompt template (\S\ref{sec:prompt_template}) to standardize model outputs and maintain consistent formatting across training stages.
Next, we perform reasoning distillation via rejection sampling to construct a high-quality dataset of biological interpretations, followed by SFT for cold-start initialization (\S\ref{sec:distill&sft}). 
Finally, we apply RL with GRPO, guided by batch-level reward signals, to further refine the model (\S\ref{sec:rl}).

\subsection{Prompt Template for Structured Reasoning}
\label{sec:prompt_template}

To ensure unified output formatting across training stages, we design a standardized prompt template that is used throughout distillation, SFT, and RL.
As shown in Table~\ref{tab:prompt_template}, this template promotes global reasoning over the entire cell batch by instructing the model to integrate contextual information and jointly assign cell types to all cells from the candidate set.

\begin{table}[h!]
    \centering
    \small
    \caption{Prompt template guiding structured reasoning and prediction in \Ours.}
    \label{tab:prompt_template}
    \vspace{2.5pt}
    \renewcommand{\arraystretch}{1.2}
    \begin{tabular}{p{13.5cm}}
        \hline
        You are an expert assistant specialized in cell type annotation. You will be given a batch of $N$ cells from the same donor, where each cell represents a unique cell type.
        For each cell, the top-expressed genes are provided in descending order of expression.
        Using both the gene expression data and donor information, determine the correct cell type for each cell.
        You will also receive a list of $N$ candidate cell types, and each candidate must be assigned to exactly one cell. Ensure that you consider all cells and candidate types together, rather than annotating each cell individually.
            Include your detailed reasoning within \thinkbegin and \thinkend tags, and provide your final answer within \answerbegin and \answerend tags.
        The final answer should be a single string listing the assigned cell types in order, separated by ` | '.
        \\
        \hline
    \end{tabular} 
\end{table}

\subsection{Reasoning Distillation and Cold Start}
\label{sec:distill&sft}

The batch-level reasoning task introduced in our benchmark poses significant challenges for model training. Unlike conventional classification tasks, the model must simultaneously analyze a group of cells, compare gene expression patterns, incorporate shared metadata, and generate a consistent label assignment. The high interdependence between inputs and outputs makes it difficult to produce valid and correct predictions, particularly at early training stages.
We begin by performing reasoning distillation using OpenAI's \texttt{o1}~\cite{jaech2024openai}, a frontier LLM with strong multi-step reasoning capabilities, to construct a synthetic dataset of high-quality reasoning traces and predictions. This distilled dataset is then used for SFT, serving as a cold start initialization~\cite{guo2025deepseek} for the subsequent RL.

\subsubsection{Reasoning Distillation via Rejection Sampling}

We use \texttt{o1}~\cite{jaech2024openai} to generate reasoning traces and predictions for 10,155 instances in \Bench. Each input is paired with the standardized prompt shown in Table~\ref{tab:prompt_template}. 
For each instance, we generate 8 candidate responses. We then apply rejection sampling to filter out low-quality outputs. A response is accepted only if it (1) adheres to the expected format and (2) yields a cell-type assignment that exactly matches the ground-truth labels. This process results in a distilled dataset containing 3,912 accepted examples, corresponding to an acceptance rate of 38.52\%. 

The collected reasoning traces exhibit several desirable properties for guiding models toward expert annotation behavior and serve as a strong foundation for SFT: First, they reflect domain-specific knowledge by referencing known gene markers and biological associations. Second, they demonstrate global reasoning behavior, in which multiple cells and candidate labels are considered jointly to ensure consistency in label assignments. Third, the reasoning is structured and interpretable, disentangled from the final prediction and formatted in a standardized prompt-response style.

\subsubsection{Supervised Fine-tuning with Distilled Reasoning}

We apply SFT on the distilled dataset to initialize the model before RL. While the final training objective involves optimizing reward signals, we find that starting from a purely pre-trained large language model fails to yield meaningful learning progress. Without prior exposure to the task-specific reasoning format and prediction structure, the model struggles to generate valid outputs. In practice, this leads to persistent formatting errors and incorrect label assignments during early-stage rollouts, resulting in zero or negative rewards and causing the policy to stagnate.

SFT serves as a cold start mechanism that mitigates this problem by providing the model with high-quality demonstrations of structured reasoning and correct predictions. By learning from the distilled dataset, the model acquires the basic capacity to follow instructions, adhere to the required response format, and reason across multiple entities within a batch. This improves response validity, reduces reward sparsity, and facilitates more stable and efficient policy optimization in the RL stage.

\subsection{Reinforcement Learning with GRPO}
\label{sec:rl}

\subsubsection{Group Relative Policy Optimization}

To reduce the training cost of RL, we adopt Group Relative Policy Optimization (GRPO)~\cite{shao2024deepseekmath}, which avoids training a separate critic by estimating advantages from a group of rollouts. 
Given a training input $x$, GRPO samples a group of $G$ candidate responses $\{y_i\}_{i=1}^{G}$ from the old policy $\pi_{\theta_{\text{old}}}$. The updated policy $\pi_\theta$ is optimized by maximizing the following objective:
\begin{equation}
\small
\begin{aligned}
\label{eq:grpo}
  \mathcal{J}(\theta) = \mathbb{E}_{x \sim \mathcal{D}, \{y_i\}_{i=1}^{G} \sim \pi_{\theta_{\text{old}}}(\cdot|x)} 
  \frac{1}{G} \sum_{i=1}^{G} \left[\min \left( p_i(\theta) A_{i}, \text{clip} \left( p_i(\theta), 1-\epsilon, 1+\epsilon \right) A_{i} \right) - \beta \mathbb{D}_{\text{KL}} \left( \pi_{\theta} || \pi_{\theta_{\text{ref}}} \right)\right]
\end{aligned}
\end{equation}
where $p_i = \frac{\pi_\theta(y_i|x)}{\pi_{\theta_{\text{old}}}(y_i|x)}$ denotes the probability ratio between the current and old policy, and the normalized advantage is computed as $A_i = \frac{r_i - \text{mean}(\{r_j\}_{j=1}^{G})}{\text{std}(\{r_j\}_{j=1}^{G})}$, where $r_i$ is the reward of the $i$-th response.
The KL divergence term $\mathbb{D}_{\text{KL}}$ penalizes deviation from a reference policy $\pi_{\theta_{\text{ref}}}$ to stabilize training. Here, $\epsilon$ is the clipping threshold, and $\beta$ controls the strength of the KL penalty.

\subsubsection{Reward Modeling}

To incentivize structured reasoning and ensure answer validity, we design a rule-based reward function that evaluates both the format and correctness of the model output. 
As instructed in Table~\ref{tab:prompt_template}, the model is required to generate responses in a strict format containing exactly one reasoning segment enclosed in \thinkbegin...\thinkend tags and one answer segment enclosed in \answerbegin...\answerend tags. Responses with incorrect tags, or extra texts, are considered invalid and receive a penalty.

For valid responses, we extract the predicted answer $\hat{\mathbf{y}} = [\hat{y}_1, \dots, \hat{y}_N]$ and compare it to the ground truth labels $\mathbf{y} = [y_1, \dots, y_N]$. The batch-level reward is computed as:
\begin{equation}
\small
R_{\text{batch}} = 
\begin{cases}
\prod_{i=1}^N \mathbf{1}(\hat{y}_i = y_i), & \text{if format is valid} \\
-1, & \text{if format is invalid}
\end{cases}
\end{equation}
Here, $\mathbf{1}(\cdot)$ is an indicator function. This setup ensures that a reward of 1 is given only when \emph{all} cell types in the batch are correctly predicted; any incorrect prediction results in a reward of 0.

\section{Experiments}

\subsection{Configuration}
\label{sec:config}
\noindent \textbf{Datasets.} 
We conduct experiments on the proposed \Bench. Each instance comprises $8 \leq N \leq 15$ cells, each represented by its ranked top-expressed genes and a shared donor-level contextual description.
For training, we perform reasoning distillation using a strong reasoning model to generate 3,912 high-quality reasoning traces, which are then used for SFT. In addition, we sample 3,000 extra instances to construct an RL dataset, resulting in a total of 6,912 instances for the GRPO stage. A held-out test set of 1,095 batches is used for final evaluation.
More details on the dataset processing can be found in the Supplementary Materials.

\noindent \textbf{Evaluation Metrics.}
We report results using the following metrics:
(1) \textit{Cell-level Accuracy:} The average proportion of correctly predicted labels per batch.
(2) \textit{Batch-level Accuracy:} The proportion of batches where all predicted labels exactly match the ground truth.
(3) \textit{Format Validity:} The proportion of outputs that follow the required response format.
(4) \textit{Answer Uniqueness:} The average proportion of unique cell type predictions per batch.

\subsection{Main Results}

Table~\ref{tab:cell_annotation_results} summarizes the overall performance of all models on the \Bench test set. We compare open-source LLMs (e.g., \texttt{Llama}~\cite{touvron2023llama}, \texttt{Qwen}~\cite{bai2023qwen}), closed-source LLMs with strong reasoning capabilities (e.g., \texttt{GPT-4o}~\cite{achiam2023gpt}, \texttt{o1}), and instruction-tuned baselines that directly predict answers from the input without producing reasoning traces. 
More details are in the Supplementary Materials.

\begin{table*}[h!]
\centering
\small
\footnotesize
\renewcommand\tabcolsep{4.5pt}
\renewcommand\arraystretch{1.05}
\caption{Evaluation of model performance on the \Bench benchmark. The best and second-best results are highlighted in \colorbox{backred!50}{red} and \colorbox{backblue!75}{blue}, respectively.}
\resizebox{0.93\textwidth}{!}{
\begin{tabular}{l|c|c|c|c|c}
\toprule
\header{Model} & \header{Response Length} & \header{Cell-level Acc} & \header{Batch-level Acc} & \header{Format} & \header{Uniqueness} \\
\midrule
\rowcolor[rgb]{0.93,0.93,0.93} \multicolumn{6}{c}{\textit{Zero-shot performance (w/ reasoning)}} \\
\texttt{Llama3.1-8B-Instruct} & 2264.9 & 0.0050 & 0.0000 & 0.0393 & 0.0190 \\
\texttt{Qwen2.5-7B-Instruct} & 879.8 & 0.1353 & 0.0000 & 0.9425 & 0.4528 \\
\texttt{Llama3.3-70B-Instruct} & 1006.8 & 0.2573	& 0.0064 & 0.9973 & 0.8321 \\
\texttt{GPT-4o-mini} & 779.9 & 0.2581 & 0.0027 & 0.9918 & 0.6565 \\
\texttt{GPT-4o} & 1087.0 & 0.4248 & 0.0283 & 0.9817 & 0.8179 \\
\texttt{o3-mini} & 1130.2 & 0.5804 & 0.1352 & \colorbox{backred!50}{1.0000} & \colorbox{backred!50}{1.0000} \\
\texttt{o1} & 906.6 & \colorbox{backblue!75}{0.6479} & 0.1900 & \colorbox{backred!50}{1.0000} & \colorbox{backblue!75}{0.9997} \\
% Cell2Sentence & -- & & & & \\
\rowcolor[rgb]{0.93,0.93,0.93} \multicolumn{6}{c}{\textit{Instruction-tuning Performance (w/o reasoning)}} \\
\texttt{Llama3.1-8B-Instruct} & 71.9 & 0.6411 & \colorbox{backblue!75}{0.2969} & 0.9078 & 0.9071 \\
\texttt{Qwen2.5-7B-Instruct} & 70.9 & 0.6179 & 0.2475 & 0.9443 & 0.9404 \\
% Cell2Sentence & -- & & & & \\
\rowcolor[rgb]{0.93,0.93,0.93} \multicolumn{6}{c}{\textit{Ours (w/ reasoning)}} \\
\Ours & 1145.6 & \colorbox{backred!50}{0.6849} & \colorbox{backred!50}{0.3288} & 0.9826 & 0.9824 \\
\bottomrule
\end{tabular}
}
\label{tab:cell_annotation_results}
\end{table*}

\Ours achieves the highest accuracy at both the cell and batch levels, outperforming all other models. Cell-level accuracy is consistently higher than batch-level accuracy across all models, which is expected given the stricter requirement of the latter.
While it trails slightly behind \texttt{o1} and \texttt{o3-mini} in format validity and answer uniqueness, \Ours still performs competitively on these dimensions, especially considering its smaller model size.
Zero-shot models like \texttt{Llama3.1-8B-Instruct} often produce overly long outputs due to repeating cell labels or generating more predictions than required. Instruction-tuned variants without reasoning generate much shorter responses with better format adherence, but lack the ability to perform explicit, interpretable reasoning.
Closed-source models such as \texttt{GPT-4o} and \texttt{o1} show stronger instruction-following and reasoning consistency, producing well-structured outputs more reliably than most open-source models.

\begin{figure}[h!]
\centering
\includegraphics[width=1.0\textwidth]{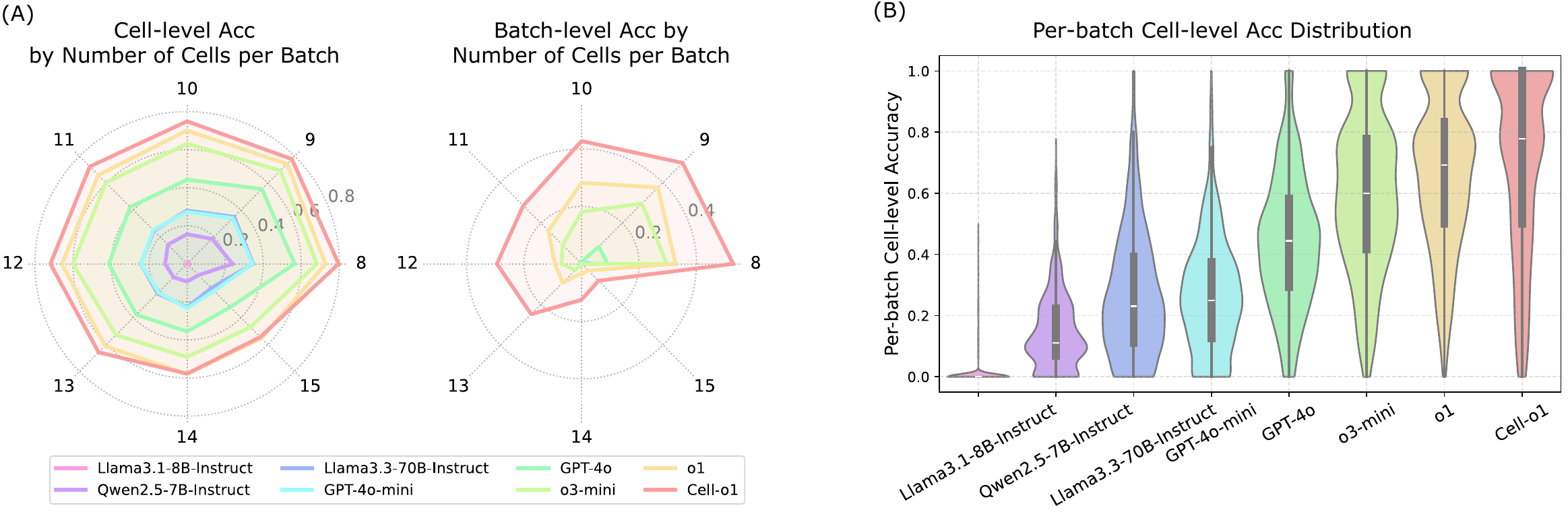}
\caption{
Fine-grained analysis of model performance.
(A) Cell-level and batch-level accuracy across varying numbers of cells per batch. 
(B) Distribution of per-batch cell-level accuracy across models.
}
\label{fig:acc_distribution}
\end{figure}

To further understand model behavior under varying conditions, we conduct a fine-grained analysis.
Figure~\ref{fig:acc_distribution} (A) shows that all models experience a decline in accuracy as the number of cells per batch increases, with \Ours consistently achieving higher scores across the full range.
Figure~\ref{fig:acc_distribution} (B) further shows that \Ours achieves the highest median and lowest variance in per-batch cell-level accuracy, indicating both strong and stable predictions across diverse contexts.

\subsection{Generalization to Unseen Diseases}

To evaluate the robustness and generalization ability of \Ours, we conduct evaluation on cell batches from disease types that were not seen during training. This setting simulates real-world scenarios where models are applied to new biological conditions without task-specific fine-tuning. 

\begin{table*}[h!]
\centering
\footnotesize
\renewcommand\tabcolsep{4pt}
\renewcommand\arraystretch{1.05}
\caption{Evaluation of model performance on unseen disease datasets.  The best and second-best results are highlighted in \colorbox{backred!50}{red} and \colorbox{backblue!75}{blue}, respectively.}
\resizebox{\textwidth}{!}{
\begin{tabular}{l|cc|cc|cc|cc|cc}
\toprule
\header{Model} & \multicolumn{2}{c|}{\textit{Breast Cancer}} & \multicolumn{2}{c|}{\textit{Melanoma}} & \multicolumn{2}{c|}{\textit{SLE}} & \multicolumn{2}{c|}{\textit{Colorectal Cancer}} & \multicolumn{2}{c}{Average} \\
 & Cell & Batch & Cell & Batch & Cell & Batch & Cell & Batch & Cell & Batch \\
\midrule
\rowcolor[rgb]{0.93,0.93,0.93} \multicolumn{11}{c}{\textit{Zero-shot (w/ reasoning)}} \\
\texttt{Llama3.1-8B-Instruct} & 0.0075 & 0.0000 & 0.0179 & 0.0000 & 0.0035 & 0.0000 & 0.0147 & 0.0000 & 0.0109 & 0.0000 \\
\texttt{Qwen2.5-7B-Instruct} & 0.2382 & 0.0000 & 0.2679 & 0.0000 &  0.2400 & 0.0000 & 0.1996 & 0.0000  & 0.2364 & 0.0000 \\
\texttt{Llama3.3-70B-Instruct} & 0.4400 & 0.0000 & 0.5337 & 0.0159 & 0.3165 & 0.0100 & 0.4218 & 0.0000 & 0.4280 & 0.0065 \\
\texttt{GPT-4o-mini} & 0.4171 & 0.0000 & 0.5198 & 0.0000 & 0.4129 & 0.0000 & 0.4549 & 0.0000 & 0.4512 & 0.0000 \\
\texttt{GPT-4o} & 0.5847 & 0.0560 & 0.7619 & 0.1587 & 0.6576 & 0.1000 & 0.5715 & 0.0239 & 0.6439 & 0.0847 \\
\texttt{o3-mini} & 0.7806 & 0.2960& 0.7341 & 0.1111 & 0.7706 & 0.3200 & 0.7208 & 0.1713 & 0.7515 & 0.2246 \\
\texttt{o1} & \colorbox{backred!50}{0.8136} & \colorbox{backblue!75}{0.3200} & 0.7738 & 0.1270 & \colorbox{backblue!75}{0.8200} & \colorbox{backblue!75}{0.3700} & \colorbox{backred!50}{0.7845} & \colorbox{backred!50}{0.3108} & \colorbox{backblue!75}{0.7980} & 0.2819 \\
\rowcolor[rgb]{0.93,0.93,0.93} \multicolumn{11}{c}{\textit{Instruction-tuned (w/o reasoning)}} \\
\texttt{Llama3.1-8B-Instruct}  & 0.7872 & 0.2880 &  \colorbox{backblue!75}{0.7837} & \colorbox{backblue!75}{0.3333} & 0.7176 & 0.2500 & 0.7208 & 0.2590 & 0.7523 & \colorbox{backblue!75}{0.2826} \\
\texttt{Qwen2.5-7B-Instruct}  & 0.7250 & 0.1600 & 0.7282 & 0.1746 & 0.5553 & 0.0300 & 0.5816 & 0.1036 & 0.6475 & 0.1170 \\
\rowcolor[rgb]{0.93,0.93,0.93} \multicolumn{11}{c}{\textit{Ours (w/ reasoning)}} \\
\Ours & \colorbox{backblue!75}{0.8089} & \colorbox{backred!50}{0.3360} & \colorbox{backred!50}{0.8591} & \colorbox{backred!50}{0.5397} & \colorbox{backred!50}{0.8235} & \colorbox{backred!50}{0.3800} & \colorbox{backblue!75}{0.7820} & \colorbox{backblue!75}{0.3028} & \colorbox{backred!50}{0.8184} & \colorbox{backred!50}{0.3896} \\
\bottomrule
\end{tabular}
}
\label{tab:generalization}
\end{table*}

Specifically, we collect raw scRNA-seq data from four disease conditions—\textit{Breast Cancer}, \textit{Melanoma}, \textit{Systemic Lupus Erythematosus (SLE)}, and \textit{Colorectal Cancer}—and follow the \Bench construction protocol to build corresponding test batches for zero-shot evaluation, resulting in a total of 539 annotated instances.
As shown in Table~\ref{tab:generalization}, \Ours consistently outperforms all baselines across both cell- and batch-level accuracy. Closed-source models such as \texttt{o1} exhibit strong generalization, but \Ours still leads by a significant margin despite its smaller model size. Compared to instruction-tuned models, which tend to make direct predictions without reasoning, \Ours generalizes more effectively to unseen disease conditions. This may be attributed to its ability to reason step-by-step during prediction, allowing it to adapt more flexibly to novel biological contexts.

\section{Analysis and Discussion}

\subsection{Training and Inference Dynamics under Different Configurations}

\begin{figure}[h!]
\centering
\includegraphics[width=1.0\textwidth]{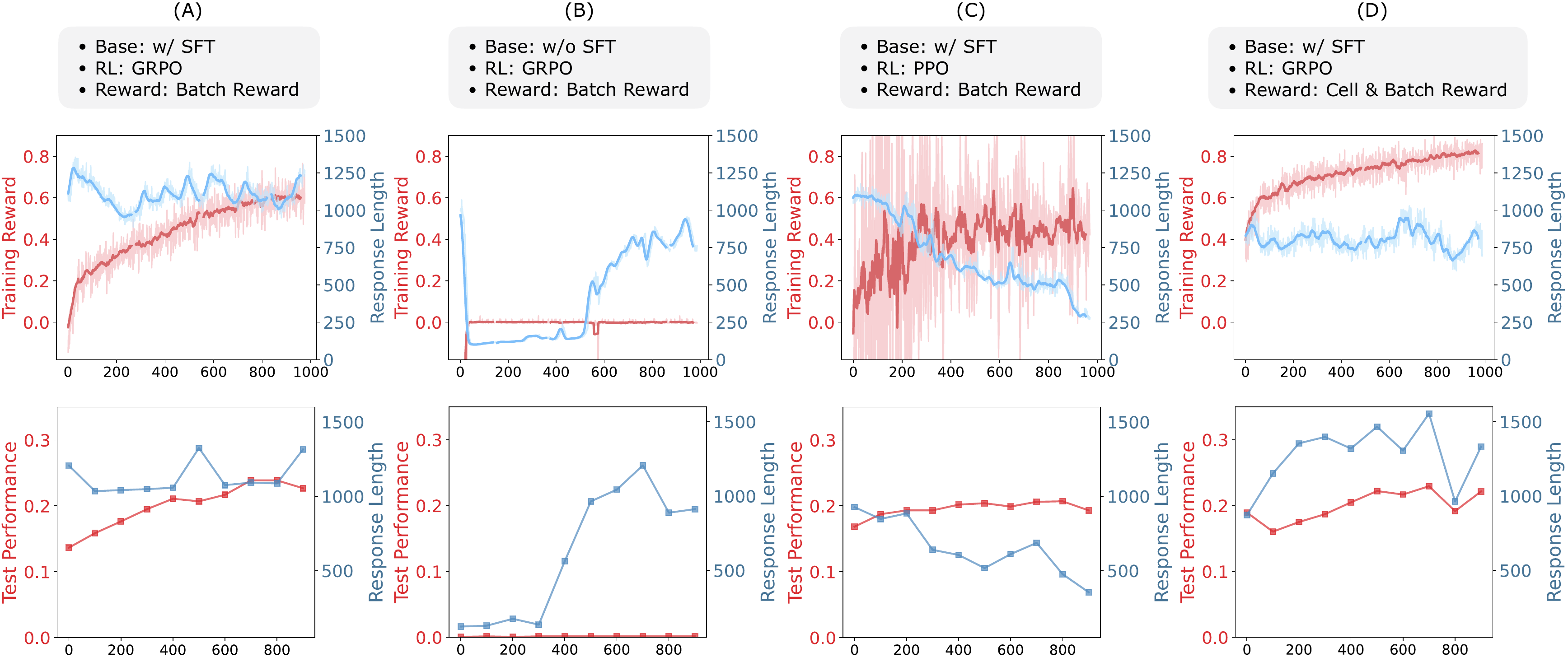}
\caption{
Comparing training and inference dynamics across different configurations.
}
\label{fig:dynamics}
\end{figure}

To understand how different training strategies affect model performance and behavior, we compare four training configurations that differ along three dimensions: the presence of SFT initialization, the RL algorithm (GRPO vs. PPO (Proximal Policy Optimization)~\cite{schulman2017proximal}), and the design of the reward function.
As shown in Figure~\ref{fig:dynamics}, \Ours (A) with SFT, GRPO, and batch-level reward achieves stable training, controlled response length, and the best test performance.
The response length fluctuates briefly at the beginning of training and then stabilizes, indicating that the model learns to balance output structure and reward optimization during training.
Removing SFT initialization (B) leads to ineffective reward learning and a sharp increase in response length, indicating poor task alignment. 
Replacing GRPO with PPO (C) results in noisier reward trajectories and limited test performance gains, likely due to a mismatch between the reward objective and the final evaluation metric.
The steadily declining response length suggests that PPO encourages shorter, less informative outputs to stabilize rewards, potentially at the expense of reasoning quality.
Configuration (D) augments the reward with a cell-level accuracy component, using the average of cell and batch-level accuracy as the training signal. This leads to higher training reward scores, but slightly less stable test performance, which remains overall comparable to configuration (A).

\subsection{Qualitative Assessment of Reasoning Traces}

To qualitatively assess the reasoning capabilities of our model, we conducted a human evaluation comparing its outputs against several baselines. We randomly sampled 100 test instances from \Bench. 
For each instance, we constructed a step-by-step prediction diagram for each model, indicating the order in which cells were annotated (denoted by numbers), the predicted label, and whether each step was correct.
If incorrect, experts further categorized the error into one of five predefined types 
%(detailed in Appendix~\ref{app_sec:evaluation}).
(detailed in Supplementary Materials).
Figure~\ref{fig:reasoning_analysis} (A) shows one example. 

To avoid bias, expert comparisons were conducted in a blind setting: for each instance, reasoning outputs from \Ours and the baseline models were anonymized and presented in a randomized order.
Experts then provided an overall judgment by comparing the reasoning quality across models, resulting in a win/tie/loss outcome for each pairwise comparison, as shown in Figure~\ref{fig:reasoning_analysis} (B).

During the human evaluation, we observed two interesting patterns exhibited by \Ours: self-reflection and curriculum reasoning.
The former refers to the model’s ability to revisit and revise its intermediate conclusions, while the latter reflects its tendency to tackle easier or more confident cases first before addressing more challenging ones. As shown in Figure~\ref{fig:reasoning_analysis} (C), \Ours is the only model that consistently demonstrates both behaviors.
This highlights its ability to structure reasoning more deliberately—similar to how human experts approach complex annotation tasks.

\begin{figure}[h!]
\centering
\includegraphics[width=1.0\textwidth]{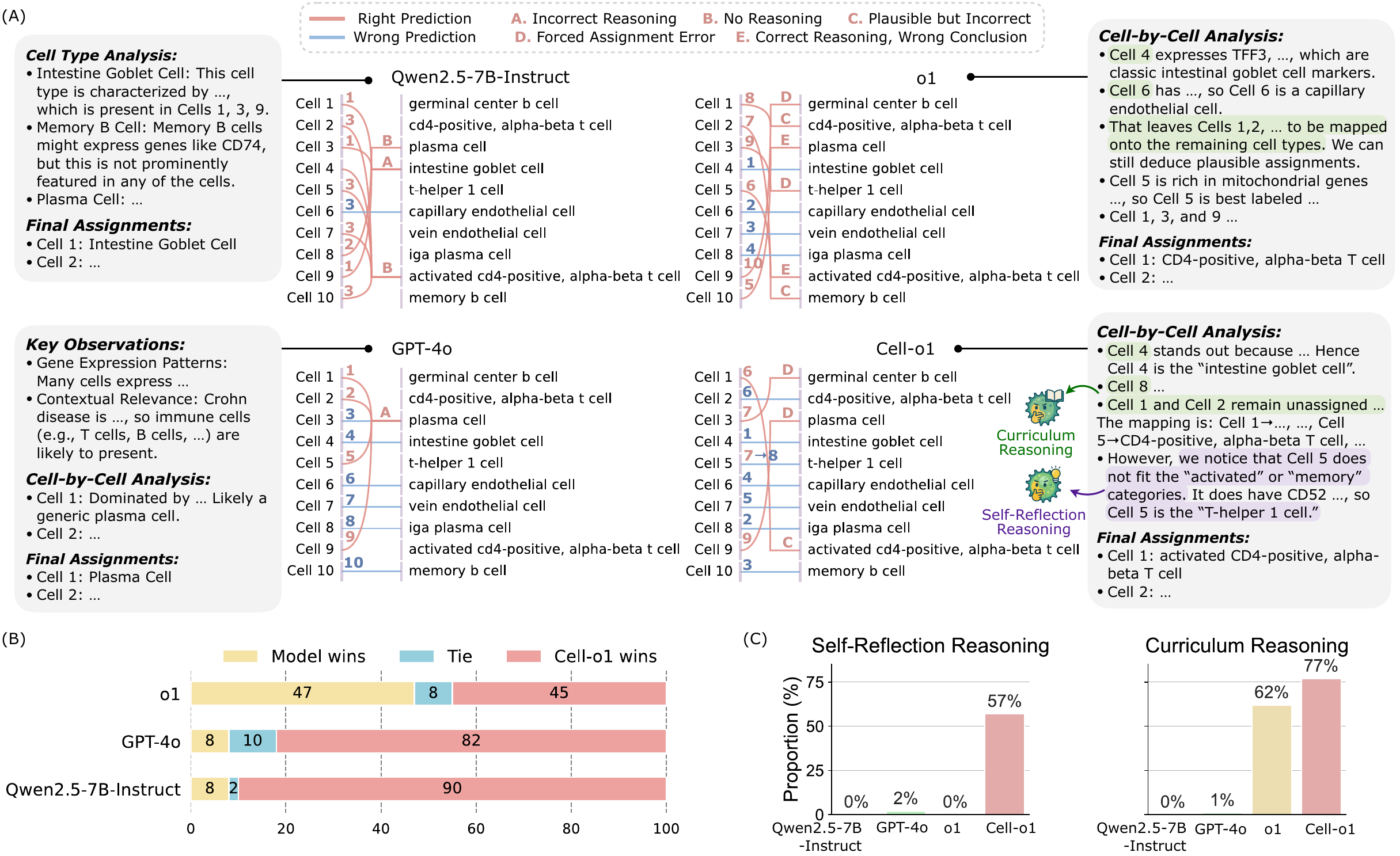}
\caption{
Analysis of model reasoning behavior. 
(A) Step-by-step reasoning traces generated by each model for the same task. 
(B) Expert evaluation of reasoning quality comparing \Ours with baseline models.
(C) Proportion of traces exhibiting self-reflection and curriculum reasoning behaviors.
}
\label{fig:reasoning_analysis}
\end{figure}

\subsection{Limitations and Broader Impacts}
While \Ours demonstrates strong batch-level reasoning for scRNA-seq cell type annotation, it relies on predefined candidate label sets, which limits its applicability to novel cell types. \Ours was trained solely on Cell\(\times\)Gene data so its generalizability to other modalities or under-represented tissues remains untested. Additionally, \Ours depends on sparse and high-variance RL rewards that require careful hyperparameter tuning, may produce plausible but incorrect reasoning, and incur non-trivial computational overhead.

Nevertheless, by providing interpretable, group-aware predictions, \Ours can accelerate annotation pipelines and democratize access to high-quality labels, which potentially fuels discoveries in immunology, oncology, and developmental biology. On the other hand, it also poses risks that misapplication could propagate errors into downstream analyses, and large-scale pre-training and fine-tuning carry environmental and economic costs, underscoring the need for responsible, human-in-the-loop deployment and careful consideration of data privacy and dual-use concerns.

\label{sec:limitation}

\section{Conclusion}

This work presents \Bench, a structured reasoning benchmark for batch-level cell type annotation designed to emulate expert workflows. 
We further introduce \Ours, an LLM that demonstrates emergent expert-like reasoning behaviors—such as self-reflection and curriculum annotation—enabling accurate and interpretable predictions across diverse cellular contexts. 
Looking ahead, future directions include incorporating retrieval-augmented methods~\cite{jin2025search,jiang2025deepretrieval,xiong2025rag,chen2025research} to integrate external knowledge, leveraging domain ontologies~\cite{hoehndorf2015role,ashburner2000gene} for more consistent reasoning~\cite{jin2024impact}, and extending structured reasoning to broader biomedical applications~\cite{tian2024opportunities}.

\section*{Acknowledgments}
This research was supported by the Division of Intramural Research (DIR) of the National Library of Medicine (NLM), National Institutes of Health.

\bibliography{references}
\bibliographystyle{unsrtnat}

\newpage

\newpage
\appendix

\begin{center}
\Large
\textbf{Appendix}
\end{center}

\section{Dataset}
\subsection{Data Sources}

Our benchmark is constructed by collecting and curating a diverse range of high-quality single-cell RNA sequencing (scRNA-seq) datasets from publicly available sources, primarily hosted on the CELLxGENE platform~\cite{czi2025cz}. These datasets cover a wide spectrum of human tissues, diseases, and biological conditions, ensuring both diversity and real-world relevance for evaluating model generalization.

Table~\ref{tab:data_source} summarizes all datasets used in our study, along with their corresponding accession links and references. The collection includes data from various immune-related diseases (e.g., rheumatoid arthritis, lupus, HIV), cancer types (e.g., lung adenocarcinoma, ovarian cancer), as well as healthy tissues across developmental and anatomical contexts. 

\begin{table*}[h]
\centering
\footnotesize
\renewcommand\tabcolsep{9pt}
\renewcommand\arraystretch{1.05}
\caption{
Raw scRNA-seq data resources involved in our paper.}
\resizebox{\textwidth}{!}{
\begin{tabular}{p{9cm} p{7.8cm}}
\toprule
\textbf{Data Source Name} & \textbf{URL} \\
\midrule
Single-cell Human Microglia~\cite{olah2020single} & \small \url{https://cellxgene.cziscience.com/collections/fcb3d1c1-03d2-41ac-8229-458e072b7a1c} \\
\midrule
Single-Cell RNA-Seq Analysis Reveals Cell Subsets And Gene Signatures Associated With Rheumatoid Arthritis Disease Activity~\cite{binvignat2024single} & \small \url{https://cellxgene.cziscience.com/collections/e1a9ca56-f2ee-435d-980a-4f49ab7a952b} \\
\midrule
The Single-Cell Lung Cancer Atlas (LuCA)~\cite{salcher2022high} & 
\small \url{https://cellxgene.cziscience.com/collections/edb893ee-4066-4128-9aec-5eb2b03f8287} \\
\midrule
Cells Of The Human Intestinal Tract Mapped Across Space And Time~\cite{elmentaite2021cells} & \small \url{https://cellxgene.cziscience.com/collections/e33ffcd3-7cbf-4b8c-b0f4-85587ad5019a} \\
\midrule
Persistent T Cell Unresponsiveness Associated With Chronic Visceral Leishmaniasis In HIV-Coinfected Patients~\cite{de2024persistent} & \small \url{https://cellxgene.cziscience.com/collections/126afc71-47fb-4e9d-8aaf-9d9f61e0ac77} \\
\midrule
Multiomics Single-Cell Analysis Of Human Pancreatic Islets Reveals Novel Cellular States In Health And Type 1 Diabetes~\cite{fasolino2022single} & \small \url{https://cellxgene.cziscience.com/collections/51544e44-293b-4c2b-8c26-560678423380} \\
\midrule
The Integrated Human Lung Cell Atlas~\cite{sikkema2023integrated} & \small \url{https://cellxgene.cziscience.com/collections/6f6d381a-7701-4781-935c-db10d30de293} \\
\midrule
Single-Cell Transcriptomics Of The Human Retinal Pigment Epithelium And Choroid In Health And Macular Degeneration~\cite{voigt2019single} & \small \url{https://cellxgene.cziscience.com/collections/f8057c47-fcd8-4fcf-88b0-e2f930080f6e} \\
\midrule
Single-Cell RNA-Seq Of The Adult Human Kidney~\cite{lake2023atlas} & \small \url{https://cellxgene.cziscience.com/collections/bcb61471-2a44-4d00-a0af-ff085512674c} \\
\midrule
MSK SPECTRUM – Ovarian Cancer Mutational Processes Drive Site-Specific Immune Evasion~\cite{vazquez2022ovarian} & \small \url{https://cellxgene.cziscience.com/collections/4796c91c-9d8f-4692-be43-347b1727f9d8} \\
\midrule
HTAN-CHOP - Single-Cell Multiomics Reveals Increased Plasticity, Resistant Populations, And Stem-Cell–Like Blasts In KMT2A-Rearranged Leukemia~\cite{chen2022single} & \small \url{https://cellxgene.cziscience.com/collections/10ec9198-584e-4a7e-8a24-4a332915a4ef} \\
\midrule
Single-Cell Atlas Of Common Variable Immunodeficiency Shows Germinal Center-Associated Epigenetic Dysregulation In B-Cell Responses~\cite{rodriguez2022single} & \small \url{https://cellxgene.cziscience.com/collections/bf325905-5e8e-42e3-933d-9a9053e9af80} \\
\midrule
Dissecting Novel Myeloid-Derived Cell States Through Single-Cell RNA-Seq And Its Impact On Clinical Outcome Across Tumor Types~\cite{guimaraes2024single} & \small \url{https://cellxgene.cziscience.com/collections/3f7c572c-cd73-4b51-a313-207c7f20f188} \\
\bottomrule
\end{tabular}
}
\label{tab:data_source}
\end{table*}

\subsection{Data Statistics}

To characterize the diversity and distribution of our curated dataset, we conduct a comprehensive analysis across three axes: disease, tissue, and cell type.

\begin{table*}[h!]
\centering
\footnotesize
\renewcommand\tabcolsep{4pt}
\renewcommand\arraystretch{1.05}
\caption{Number of samples per disease in the dataset.}
\resizebox{0.9\textwidth}{!}{
\begin{tabular}{l r @{\hskip 1.5cm} l r}
\toprule
\textbf{Disease} & \textbf{\#Samples} & \textbf{Disease} & \textbf{\#Samples} \\
\midrule
Lung Adenocarcinoma & 1292 & Crohn Disease & 215 \\
Chronic Obstructive Pulmonary Disease & 1194 & Lung Large Cell Carcinoma & 211 \\
Squamous Cell Lung Carcinoma & 946 & Non-Specific Interstitial Pneumonia & 172 \\
Normal & 941 & Chronic Rhinitis & 158 \\
Interstitial Lung Disease & 777 & Hypersensitivity Pneumonitis & 126 \\
Pulmonary Fibrosis & 669 & Lymphangioleiomyomatosis & 109 \\
Non-Small Cell Lung Carcinoma & 632 & Pleomorphic Carcinoma & 87 \\
B-Cell Acute Lymphoblastic Leukemia & 74 & Pulmonary Sarcoidosis & 58 \\
Pneumonia & 48 & Acute Myeloid Leukemia & 45 \\
Common Variable Immunodeficiency & 45 & Type 1 Diabetes Mellitus & 35 \\
Acute Promyelocytic Leukemia & 28 & Rheumatoid Arthritis & 23 \\
Acute Kidney Failure & 20 & Malignant Ovarian Serous Tumor & 18 \\
Chronic Kidney Disease & 18 & Age Related Macular Degeneration 7 & 13 \\
Basal Laminar Drusen & 12 & Cystic Fibrosis & 11 \\
Hiv Infectious Disease & 11 & Alzheimer Disease & 8 \\
Cataract & 7 & Temporal Lobe Epilepsy & 4 \\
\bottomrule
\end{tabular}
}
\label{tab:disease_stats}
\end{table*}

Table~\ref{tab:disease_stats} summarizes the number of samples for each disease. The dataset spans a wide range of pathological conditions, including both common and rare diseases, such as lung adenocarcinoma, rheumatoid arthritis, HIV infection, and macular degeneration.

\begin{table}[ht]
\centering
\footnotesize
\renewcommand\tabcolsep{4pt}
\renewcommand\arraystretch{1.05}
\caption{Number of unique cell types per tissue in the dataset.}
\vskip 0.06in
\resizebox{0.75\textwidth}{!}{
\begin{tabular}{l r @{\hskip 1.5cm} l r}
\toprule
\textbf{Tissue} & \textbf{\#Cell Types} & \textbf{Tissue} & \textbf{\#Cell Types} \\
\midrule
Ileum & 71 & Adrenal Tissue & 18 \\
Lung & 56 & Rectum & 18 \\
Blood & 48 & Ascending Colon & 15 \\
Duodenum & 48 & Mesenteric Lymph Node & 11 \\
Colon & 40 & Peripheral Region Of Retina & 10 \\
Bone Marrow & 38 & Fovea Centralis & 10 \\
Nose & 36 & Right Ovary & 9 \\
Sigmoid Colon & 35 & Peritoneum & 9 \\
Vermiform Appendix & 33 & Islet Of Langerhans & 9 \\
Transverse Colon & 31 & Omentum & 9 \\
Descending Colon & 31 & Ascitic Fluid & 9 \\
Lymph Node & 30 & Left Ovary & 9 \\
Brain & 27 & Adnexa Of Uterus & 9 \\
Caecum & 27 & Diaphragm & 8 \\
Jejunum & 27 & Small Intestine & 8 \\
Kidney & 26 & Abdominal Wall & 8 \\
Pleural Effusion & 23 & Temporal Cortex & 6 \\
Liver & 21 & Dorsolateral Prefrontal Cortex & 5 \\
\bottomrule
\end{tabular}
}
\label{tab:tissue_stats}
\end{table}

Table~\ref{tab:tissue_stats} reports the number of unique cell types observed in each tissue. Tissues such as ileum, lung, and blood show high diversity, indicating rich intra-tissue heterogeneity for reasoning tasks.

\begin{table}[ht]
\centering
\footnotesize
\renewcommand\tabcolsep{4pt}
\renewcommand\arraystretch{1.05}
\caption{Frequency of cell types in the dataset (Part I).}
\vskip 0.06in
\resizebox{\textwidth}{!}{
\begin{tabular}{p{6cm} r p{6cm} r}
\toprule
\textbf{Cell Type} & \textbf{\#Instances} & \textbf{Cell Type} & \textbf{\#Instances} \\
\midrule
CD4-Positive, Alpha-Beta T Cell & 4428  & Fibroblast & 72  \\
CD8-Positive, Alpha-Beta T Cell & 4276  & Mucosal Invariant T Cell & 70  \\
Classical Monocyte & 4167  & Precursor B Cell & 70  \\
Natural Killer Cell & 4097  & Glial Cell & 68  \\
Alveolar Macrophage & 4051  & Mature B Cell & 68  \\
Cd1C-Positive Myeloid Dendritic Cell & 3965  & Pro-B Cell & 68  \\
B Cell & 3495  & Stem Cell & 67  \\
Pulmonary Alveolar Type 2 Cell & 3395  & Early Pro-B Cell & 66  \\
Mast Cell & 3062  & Stromal Cell Of Lamina Propria Of Small Intestine & 63  \\
Plasma Cell & 2939  & Common Lymphoid Progenitor & 59  \\
Vein Endothelial Cell & 2659  & Immature B Cell & 58  \\
Capillary Endothelial Cell & 2328  & Common Dendritic Progenitor & 57  \\
Non-Classical Monocyte & 2120  & Megakaryocyte-Erythroid Progenitor Cell & 56  \\
Macrophage & 2070  & Lung Neuroendocrine Cell & 54  \\
Pulmonary Alveolar Type 1 Cell & 1895  & Granulocyte Monocyte Progenitor Cell & 50  \\
Ciliated Columnar Cell Of Tracheobronchial Tree & 1734  & Reticular Cell & 48  \\
Elicited Macrophage & 1726  & Effector Memory Cd4-Positive, Alpha-Beta T Cell & 48  \\
Regulatory T Cell & 1490  & Endothelial Cell Of Artery & 45  \\
Malignant Cell & 1363  & Kidney Collecting Duct Principal Cell & 42  \\
Endothelial Cell Of Lymphatic Vessel & 1362  & Effector Memory Cd8-Positive, Alpha-Beta T Cell & 42  \\
Epithelial Cell Of Lung & 1272  & Epithelial Cell Of Proximal Tubule & 42  \\
Multi-Ciliated Epithelial Cell & 1124  & Myeloid Dendritic Cell & 42  \\
Bronchus Fibroblast Of Lung & 1123  & Kidney Loop Of Henle Thick Ascending Limb Epithelial Cell & 41  \\
Pulmonary Artery Endothelial Cell & 1086  & Central Memory Cd4-Positive, Alpha-Beta T Cell & 41  \\
Plasmacytoid Dendritic Cell & 948  & Epithelial Cell & 40  \\
Fibroblast Of Lung & 909  & Ionocyte & 39  \\
Respiratory Basal Cell & 897  & Effector Memory Cd8-Positive, Alpha-Beta T Cell, Terminally Differentiated & 39  \\
Tracheobronchial Smooth Muscle Cell & 857  & Late Promyelocyte & 39  \\
Alveolar Type 1 Fibroblast Cell & 844  & Kidney Interstitial Cell & 39  \\
Myeloid Cell & 806  & T-Helper 17 Cell & 38  \\
Alveolar Adventitial Fibroblast & 722  & Early Promyelocyte & 38  \\
\bottomrule
\end{tabular}
}
\label{tab:celltype_stats_part1}
\end{table}

\begin{table}[ht]
\centering
\footnotesize
\renewcommand\tabcolsep{4pt}
\renewcommand\arraystretch{1.05}
\caption{Frequency of cell types in the dataset (Part II).}
\vskip 0.06in
\resizebox{\textwidth}{!}{
\begin{tabular}{p{6cm} r p{6cm} r}
\toprule
\textbf{Cell Type} & \textbf{\#Instances} & \textbf{Cell Type} & \textbf{\#Instances} \\
\midrule
Epithelial Cell Of Lower Respiratory Tract & 702  & Germinal Center B Cell & 38  \\
Conventional Dendritic Cell & 701  & Class Switched Memory B Cell & 37  \\
Club Cell & 685  & Kidney Connecting Tubule Epithelial Cell & 36  \\
Dendritic Cell & 681  & Hematopoietic Stem Cell & 36  \\
Lung Pericyte & 569  & Unswitched Memory B Cell & 36  \\
Smooth Muscle Cell & 464  & Mononuclear Phagocyte & 36  \\
T Cell & 463  & Kidney Collecting Duct Intercalated Cell & 34  \\
Nasal Mucosa Goblet Cell & 447  & Kidney Distal Convoluted Tubule Epithelial Cell & 33  \\
Pericyte & 439  & Lymphoid Lineage Restricted Progenitor Cell & 33  \\
Lung Macrophage & 422  & Plasmablast & 33  \\
Mesothelial Cell & 412  & Inflammatory Macrophage & 32  \\
Neutrophil & 397  & Parietal Epithelial Cell & 31  \\
Stromal Cell & 264  & Myelocyte & 31  \\
Naive B Cell & 242  & Kidney Loop Of Henle Thin Ascending Limb Epithelial Cell & 30  \\
Memory B Cell & 235  & Bronchial Goblet Cell & 30  \\
Myofibroblast Cell & 222  & Tracheobronchial Serous Cell & 30  \\
Cd8-Positive, Alpha-Beta Memory T Cell & 216  & T-Helper 1 Cell & 30  \\
Endothelial Cell & 210  & Melanocyte & 29  \\
Monocyte & 191  & Schwann Cell & 29  \\
Gamma-Delta T Cell & 174  & Retinal Pigment Epithelial Cell & 28  \\
Naive Thymus-Derived Cd4-Positive, Alpha-Beta T Cell & 140  & Kidney Loop Of Henle Thin Descending Limb Epithelial Cell & 26  \\
Activated Cd8-Positive, Alpha-Beta T Cell & 126  & Erythroid Progenitor Cell & 26  \\
Iga Plasma Cell & 125  & Intestinal Tuft Cell & 25  \\
Pancreatic A Cell & 123  & Central Memory Cd8-Positive, Alpha-Beta T Cell & 25  \\
Pancreatic Ductal Cell & 123  & Kidney Interstitial Alternatively Activated Macrophage & 25  \\
Pancreatic D Cell & 122  & Basophil Mast Progenitor Cell & 24  \\
Pancreatic Stellate Cell & 122  & Serous Secreting Cell & 23  \\
Pp Cell & 122  & Cd4-Positive, Alpha-Beta Cytotoxic T Cell & 22  \\
Pancreatic Acinar Cell & 122  & Colon Epithelial Cell & 22  \\
Naive Thymus-Derived Cd8-Positive, Alpha-Beta T Cell & 121  & Cytotoxic T Cell & 21  \\
Type B Pancreatic Cell & 119  & Acinar Cell & 21  \\
Mature Nk T Cell & 119  & Type Ec Enteroendocrine Cell & 20  \\
Cd16-Positive, Cd56-Dim Natural Killer Cell, Human & 118  & Erythroid Lineage Cell & 19  \\
Cd4-Positive, Alpha-Beta Memory T Cell & 117  & Fallopian Tube Secretory Epithelial Cell & 18  \\
Activated Cd4-Positive, Alpha-Beta T Cell & 116  & M Cell Of Gut & 18  \\
Transit Amplifying Cell & 104  & Mucus Secreting Cell & 15  \\
Intestine Goblet Cell & 102  & Megakaryocyte Progenitor Cell & 15  \\
Brush Cell Of Tracheobronchial Tree & 96  & Mesenchymal Lymphangioblast & 13  \\
Igg Plasma Cell & 94  & Non-Terminally Differentiated Cell & 13  \\
Cd16-Negative, Cd56-Bright Natural Killer Cell, Human & 86  & Podocyte & 13  \\
Pancreatic Epsilon Cell & 85  & Cd14-Positive Monocyte & 12  \\
Epithelial Cell Of Alveolus Of Lung & 84  & Plasmacytoid Dendritic Cell, Human & 12  \\
Respiratory Hillock Cell & 83  & Enteroendocrine Cell & 12  \\
Enterocyte Of Colon & 79  & Microglial Cell & 12  \\
T Follicular Helper Cell & 75  & Group 3 Innate Lymphoid Cell & 11  \\
Hematopoietic Multipotent Progenitor Cell & 75  & Erythrocyte & 10  \\
\bottomrule
\end{tabular}
}
\label{tab:celltype_stats_part2}
\end{table}

Table~\ref{tab:celltype_stats_part1} and \ref{tab:celltype_stats_part2} summarize the frequency distribution of all cell types in the dataset. The distribution follows a long-tail pattern: a small number of immune-related cell types (e.g., CD4+ T cells, monocytes, macrophages) appear frequently, while many others are rare or highly tissue-specific. This distribution reflects the biological heterogeneity of real-world data and underscores the importance of building generalizable and context-aware models capable of handling diverse cellular identities~\cite{zhai2024distribution, zhao2025celler}.

\subsection{Task Definition}

We define the task as batch-level cell type assignment with contextual reasoning. Each input instance consists of a batch of $N$ cells sampled from the same donor, along with shared metadata (e.g., tissue, disease, sex, age) and a set of top expressed genes extracted from the gene expression profile of each cell. The model is also provided with $N$ candidate cell type labels. The goal is to assign each cell to a unique label, resulting in a one-to-one matching between cells and types.

This formulation differs from traditional LLM-based single-cell annotation, which typically treats each cell in isolation. By contrast, our task requires the model to reason jointly across the batch: decisions for one cell may depend on contextual clues or comparisons with other cells. This setup reflects the process followed by human experts, who often annotate cells by integrating expression trends, donor-level information, and biological prior knowledge.

The task also introduces combinatorial ambiguity, especially when the candidate cell types are biologically similar (e.g., naïve vs. memory T cells) or when the expression differences across cells are subtle. Effective solutions must leverage both global context and cell-level signals to make consistent, interpretable assignments.

\subsection{Data Construction}

\begin{table*}[h]
\centering
\footnotesize
\renewcommand\tabcolsep{4pt}
\renewcommand\arraystretch{1.05}
\caption{
Templates for converting structured donor metadata into context descriptions.}
\resizebox{0.9\textwidth}{!}{
\begin{tabular}{l p{10.5cm}}
\toprule
\textbf{Feature} & \textbf{Template} \\
\midrule
\raisebox{-.25\height}{\includegraphics[height=1.5em]{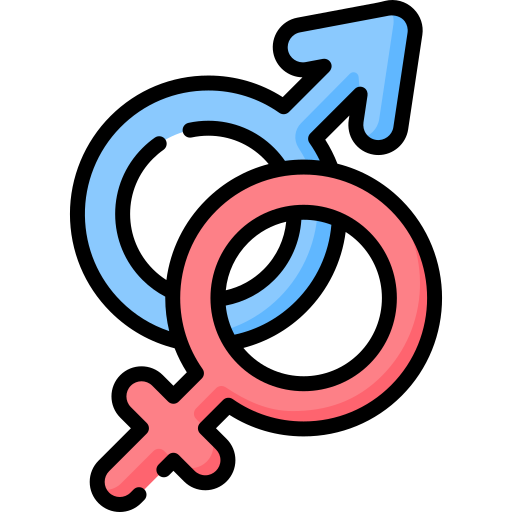}}\; Sex &
\texttt{\{sex\}} $\rightarrow$ The cell is from a \texttt{\{sex\}} individual. \\
\raisebox{-.25\height}{\includegraphics[height=1.5em]{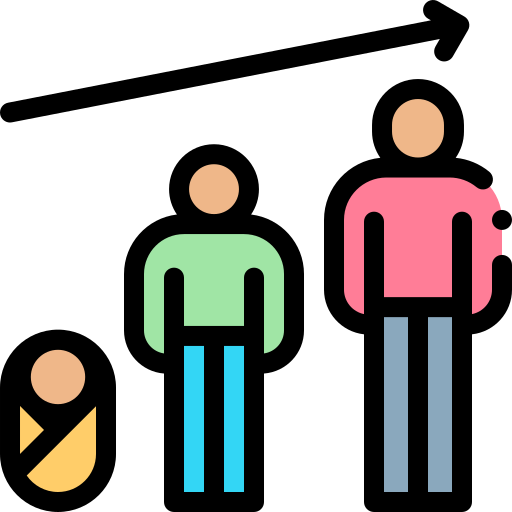}}\; Development Stage & The individual is at the \{\texttt{development\_stage}\}. \\
\raisebox{-.25\height}{\includegraphics[height=1.5em]{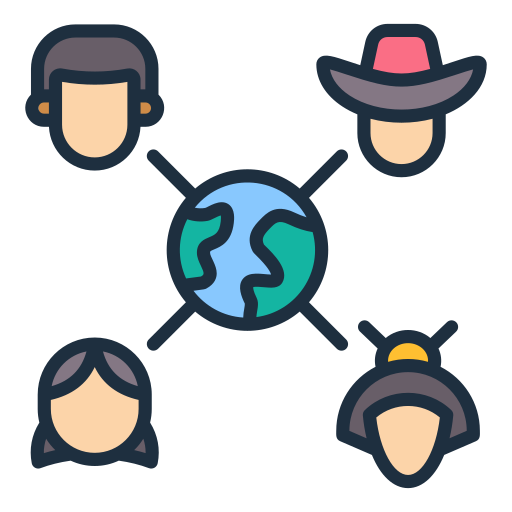}}\; Ethnicity & The donor has a \{\texttt{self\_reported\_ethnicity}\} background. \\
\raisebox{-.25\height}{\includegraphics[height=1.5em]{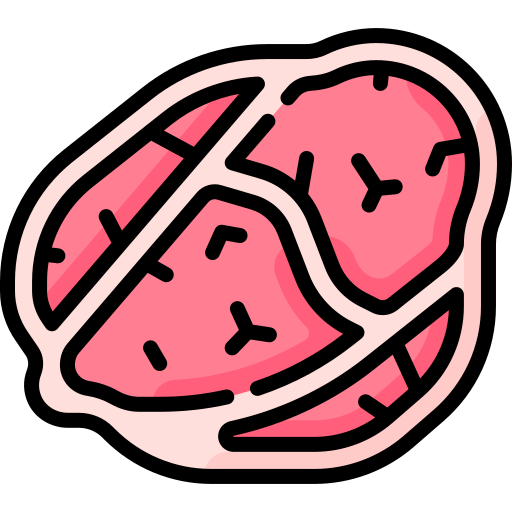}}\; Tissue & The cell originates from the \{\texttt{tissue}\}. \\
\raisebox{-.25\height}{\includegraphics[height=1.5em]{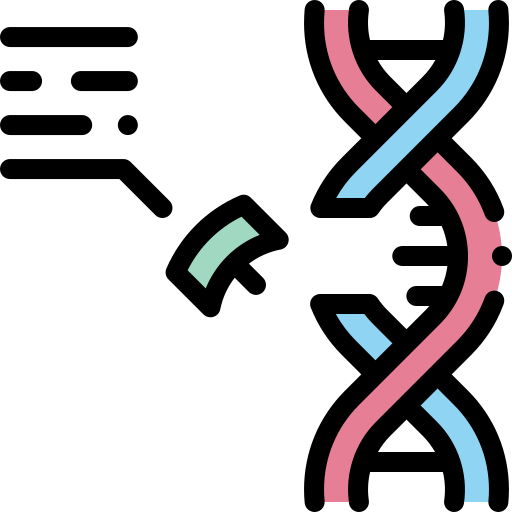}}\; Top Genes & \{\texttt{genes}\} → Top expressed genes are: \{\texttt{genes}\}. \\
\raisebox{-.25\height}{\includegraphics[height=1.5em]{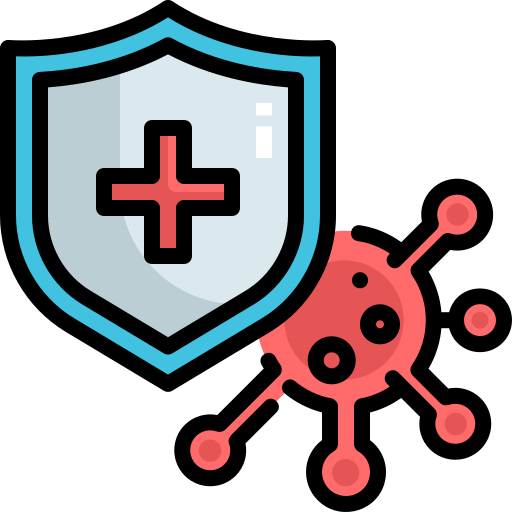}}\; Disease & 
\texttt{normal} → The patient is healthy with \texttt{no diagnosed disease}. |
\texttt{other} → The patient has been diagnosed with \{\texttt{disease}\}. \\
\raisebox{-.25\height}{\includegraphics[height=1.5em]{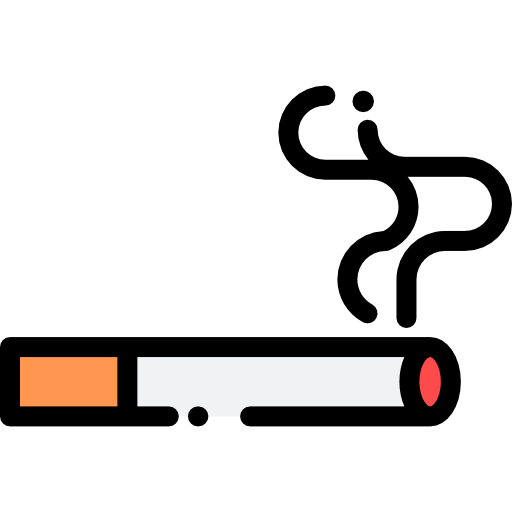}}\; Smoking Status & 
\texttt{active} → The patient is an \texttt{active} smoker. |
\texttt{former} → The patient is a \texttt{former} smoker. |
\texttt{hist of marijuana use} → The patient has a \texttt{history of marijuana use}; 
\texttt{never} → The patient has \texttt{never} smoked. \\
\raisebox{-.25\height}{\includegraphics[height=1.5em]{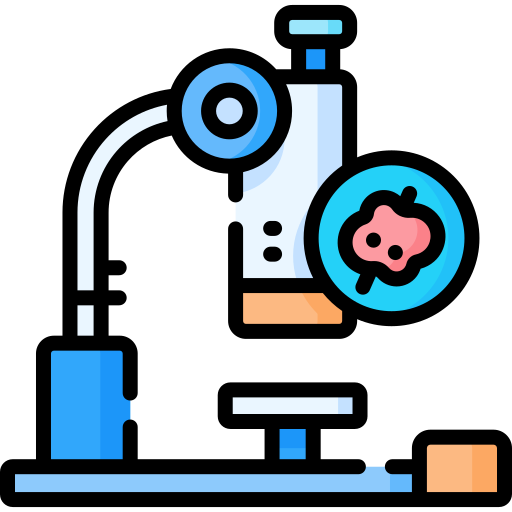}}\; Tumor Stage & 
\texttt{non-cancer} $\rightarrow$ There is \texttt{no cancer} present. |
\texttt{early} $\rightarrow$ The patient has an \texttt{early}-stage tumor. |
\texttt{advanced} $\rightarrow$ The patient has an \texttt{advanced}-stage tumor. \\
\raisebox{-.25\height}{\includegraphics[height=1.5em]{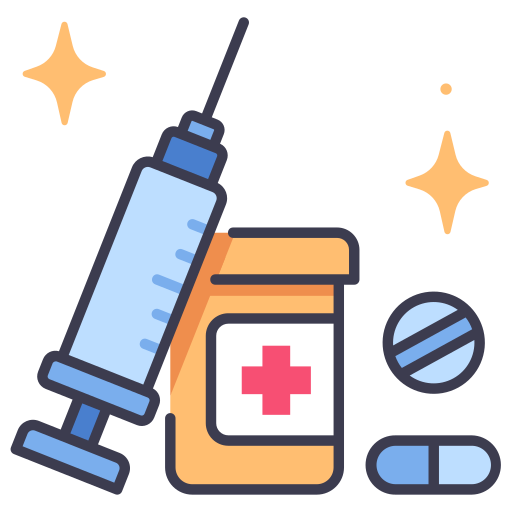}}\; Sample Type & 
\texttt{M3} $\rightarrow$ The sample was collected at \texttt{month 3} post-treatment. |
\texttt{M6} $\rightarrow$ The sample was collected at \texttt{month 6} post-treatment. |
\texttt{UV} $\rightarrow$ The sample was exposed to \texttt{ultraviolet (UV)} treatment. |
\texttt{CONTROL} $\rightarrow$ The sample is from the \texttt{control} group. \\
\raisebox{-.25\height}{\includegraphics[height=1.5em]{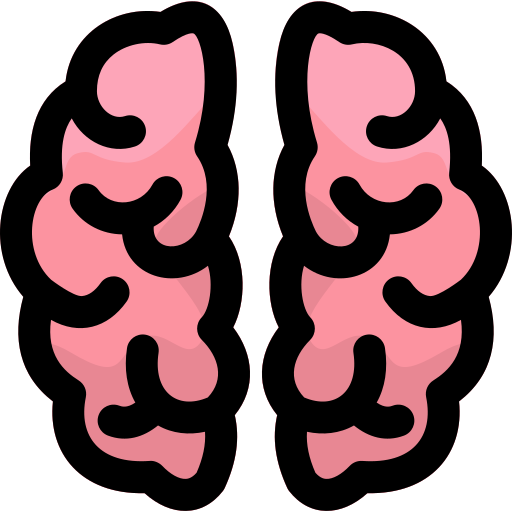}}\; Hemisphere & 
\texttt{left} $\rightarrow$ The tissue sample was taken from the \texttt{left} hemisphere of the brain. |
\texttt{right} $\rightarrow$ The tissue sample was taken from the \texttt{right} hemisphere of the brain. \\
\raisebox{-.25\height}{\includegraphics[height=1.5em]{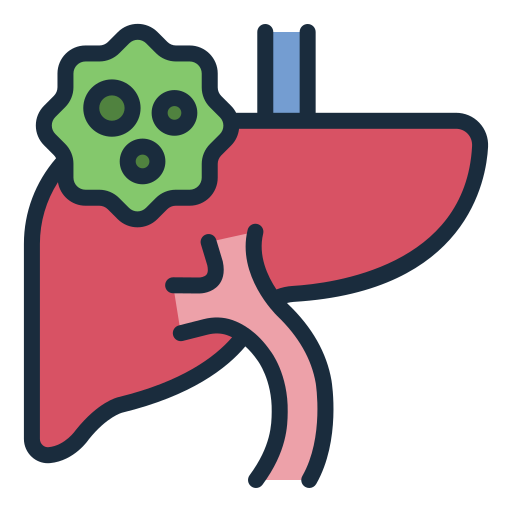}}\;  Tumor Site & 
\texttt{primary} $\rightarrow$ The tumor is located at the \texttt{primary} site. |
\texttt{metastasis} $\rightarrow$ The tumor has \texttt{metastasized} to other parts of the body. |
\texttt{normal} $\rightarrow$ This sample was collected from non-tumorous tissue. \\
\raisebox{-.25\height}{\includegraphics[height=1.5em]{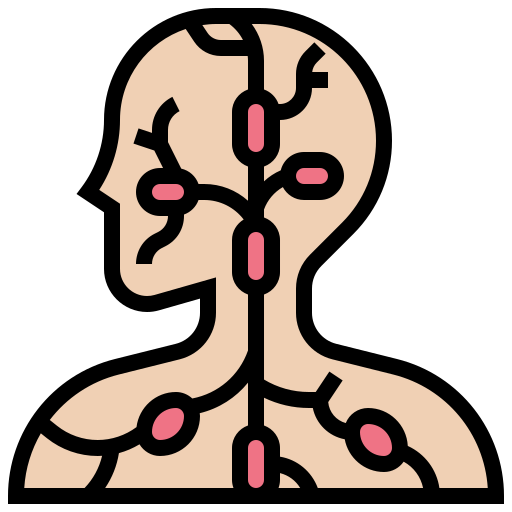}}\; Sample Source & 
\texttt{tumor} $\rightarrow$ The sample is derived from \texttt{tumor} tissue. |
\texttt{normal} $\rightarrow$ The sample is derived from \texttt{normal} tissue. |
\texttt{blood} $\rightarrow$ The sample is a \texttt{blood}-derived specimen. |
\texttt{lymphnode} $\rightarrow$ The sample is derived from \texttt{lymph node} tissue. \\
\raisebox{-.25\height}{\includegraphics[height=1.5em]{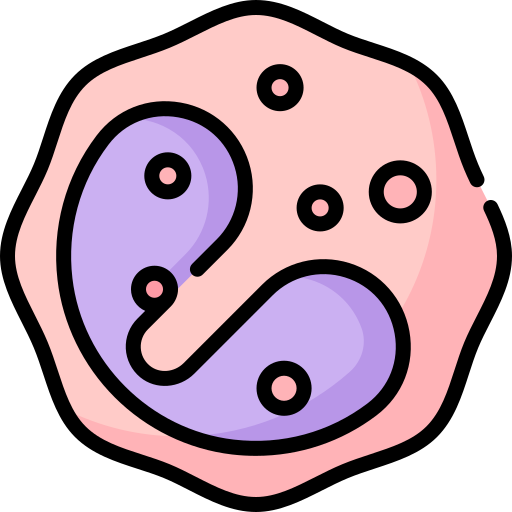}}\;  CD45 Expression & 
\texttt{yes} $\rightarrow$ The cell is CD45-\texttt{positive}, suggesting an immune cell origin. |
\texttt{no} $\rightarrow$ The cell is CD45-\texttt{negative}, suggesting a non-immune cell lineage. |
\texttt{mixed} $\rightarrow$ The sample contains a \texttt{mixture} of CD45-positive and CD45-negative cells. \\
\raisebox{-.25\height}{\includegraphics[height=1.5em]{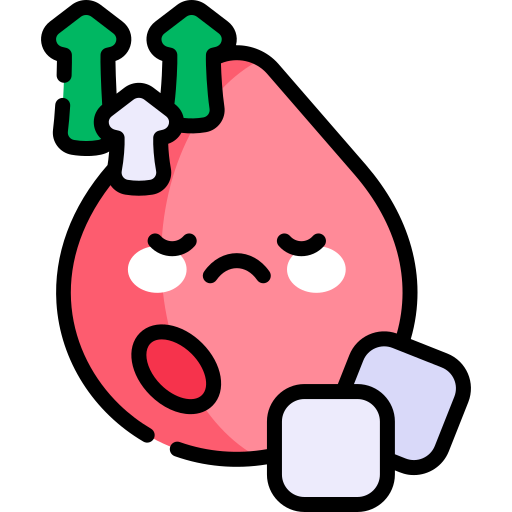}}\;  Diabetes History & 
\texttt{yes} $\rightarrow$ The patient \texttt{has a history} of diabetes. |
\texttt{no} $\rightarrow$ The patient \texttt{does not have} diabetes. \\
\raisebox{-.25\height}{\includegraphics[height=1.5em]{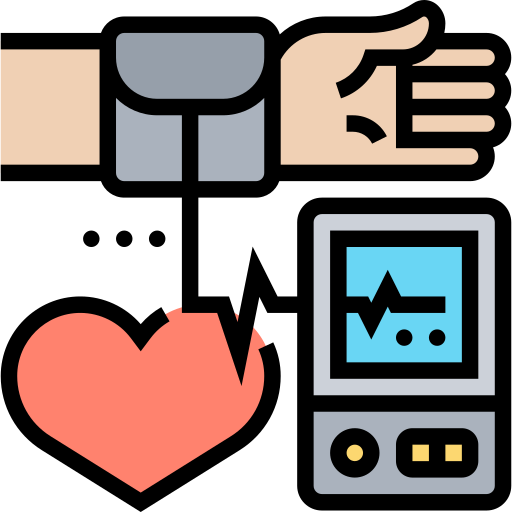}}\;  Hypertension & 
\texttt{yes} $\rightarrow$ The patient \texttt{has a history} of hypertension. |
\texttt{no} $\rightarrow$ The patient \texttt{does not have} hypertension. \\
\raisebox{-.25\height}{\includegraphics[height=1.5em]{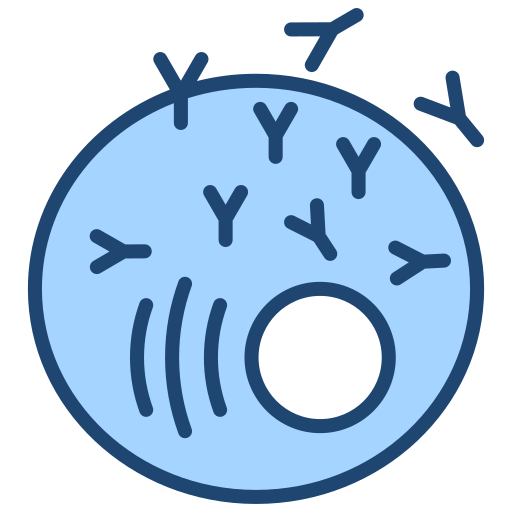}}\;  Activation & 
\texttt{activated} $\rightarrow$ The sample was stimulated and represents \texttt{activated} immune cells. |
\texttt{resting} $\rightarrow$ The sample represents \texttt{resting} (non-activated) immune cells. \\
\raisebox{-.25\height}{\includegraphics[height=1.5em]{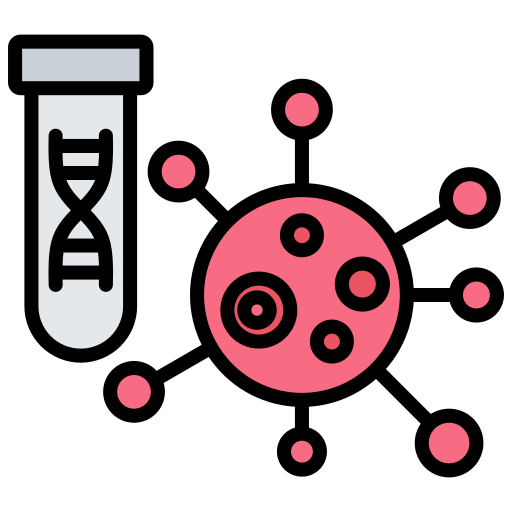}}\; Genotype & 
\texttt{FLT3-ITD,NPM1-MUT} $\rightarrow$ The patient carries FLT3-ITD and NPM1 mutations. |
\texttt{FLT3-WT,NPM1-MUT} $\rightarrow$ The patient carries a wild-type FLT3 and an NPM1 mutation. |
\texttt{APL} $\rightarrow$ The patient is diagnosed with acute promyelocytic leukemia (APL). \\
\raisebox{-.25\height}{\includegraphics[height=1.5em]{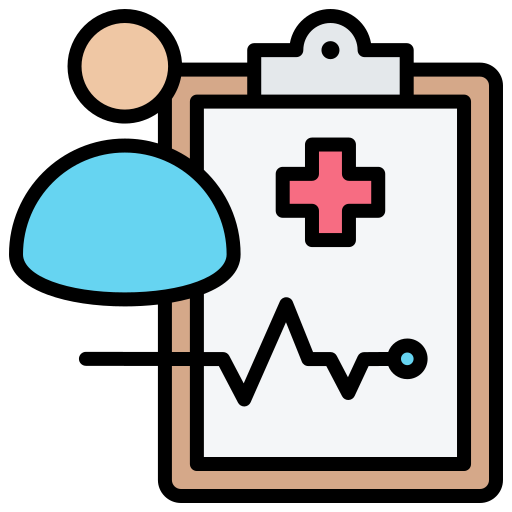}}\; Medical Conditions & 
\texttt{epilepsy} $\rightarrow$ The patient has a history of \texttt{epilepsy}. |
\texttt{tumor} $\rightarrow$ The patient has a diagnosed brain \texttt{tumor}. |
\texttt{hydrocephalus} $\rightarrow$ The patient has \texttt{hydrocephalus} (fluid buildup in the brain). |
\texttt{both} $\rightarrow$ The patient has \texttt{both} epilepsy and a brain tumor. |
\texttt{other} $\rightarrow$ The patient has \texttt{other} neurological conditions. |
\texttt{healthy} $\rightarrow$ The donor was \texttt{healthy} with no reported skin condition. |
\texttt{dm – non ulcer} $\rightarrow$ The donor had diabetes \texttt{mellitus without skin ulceration}. | 
\texttt{keloid} $\rightarrow$ The donor had a \texttt{keloid}, which is an overgrowth of scar tissue. |
\texttt{localised scleroderma} $\rightarrow$ The donor was diagnosed with \texttt{localized scleroderma}. | 
\texttt{scar} $\rightarrow$ The donor had a typical \texttt{scar} from prior skin injury. |
\texttt{\{lung\_condition\}} $\rightarrow$ The lung condition is described as \texttt{\{lung\_condition\}}. \\
\raisebox{-.25\height}{\includegraphics[height=1.5em]{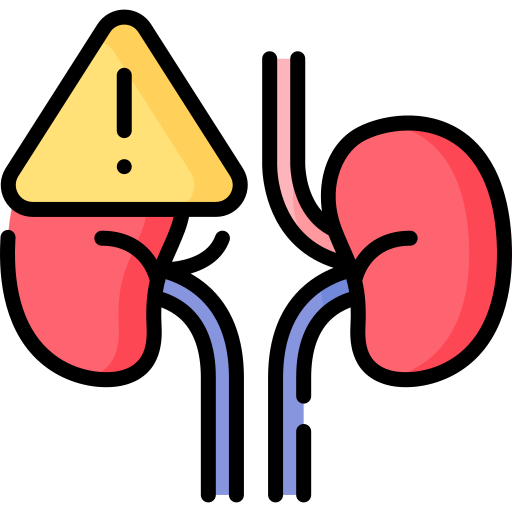}}\; Egfr & \texttt{\{value\}} $\rightarrow$ The patient's estimated glomerular filtration rate (eGFR) is in the range \texttt{\{value\}}. \\
\bottomrule
\end{tabular}
}
% \vskip -0.1in
\label{tab:template}
\end{table*}

\begin{figure}[h!]
\centering
\includegraphics[width=1.0\textwidth]{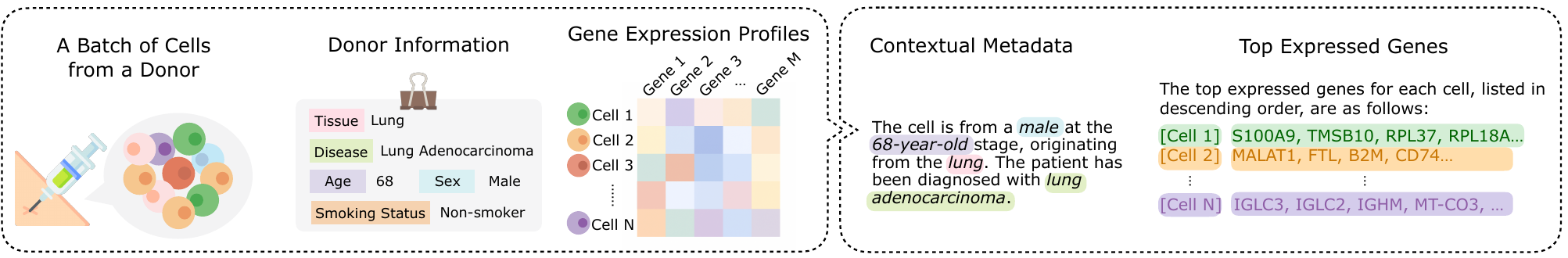}
\caption{
Overview of data construction. Each batch consists of $N$ cells from the same donor, with associated metadata (e.g., tissue, disease, age, sex, smoking status) and gene expression profiles. Structured donor-level information is converted into natural language context, and cell-level features are enriched with the top expressed genes to support downstream reasoning.
}
\label{fig:data_construction}
\end{figure}

To support the proposed task, we construct a benchmark dataset, \Bench, by assembling donor-level metadata, gene expression profiles, and curated cell type annotations from publicly available scRNA-seq datasets. 

Figure~\ref{fig:data_construction} illustrates the data construction pipeline. For each cell, we convert structured metadata fields into natural language context using predefined templates (see Table~\ref{tab:template}). These templates cover a wide range of attributes, including sex, developmental stage, tissue, disease status, and immune history. When available, we also include molecular and histological details such as tumor stage, genotype, and expression markers. The templates aim to standardize the expression of metadata fields in natural language, ensuring consistency and interpretability across different input instances.

\section{Experimental Setup Details}

\subsection{Evaluation Metrics}
\label{app:metric_formulas}

We provide the formal definitions of the evaluation metrics used in our experiments.

\begin{itemize}[leftmargin=*]
    \item \textbf{Cell-level Accuracy:}
    \[
    \text{Cell Acc} = \frac{1}{|\mathcal{D}|} \sum_{j=1}^{|\mathcal{D}|} \left( \frac{1}{N_j} \sum_{i=1}^{N_j} \mathbf{1}(\hat{y}^{(j)}_i = y^{(j)}_i) \right),
    \]
    where \( \mathcal{D} \) is the set of evaluation batches, \( N_j \) is the number of cells in batch \( j \), and \( \hat{y}^{(j)}_i \) and \( y^{(j)}_i \) denote the predicted and ground-truth labels of the \( i \)-th cell in batch \( j \).

    \item \textbf{Batch-level Accuracy:}
    \[
    \text{Batch Acc} = \frac{1}{|\mathcal{D}|} \sum_{j=1}^{|\mathcal{D}|} \prod_{i=1}^{N_j} \mathbf{1}(\hat{y}^{(j)}_i = y^{(j)}_i),
    \]
    A value of 1 is counted only when all labels in a batch are correct.

    \item \textbf{Format Validity:}
    \[
    \text{Format Validity} = \frac{1}{|\mathcal{D}|} \sum_{j=1}^{|\mathcal{D}|} \mathbf{1}\left( \hat{\mathbf{y}}^{(j)} \in \mathcal{F} \right),
    \]
    where $\hat{\mathbf{y}}^{(j)}$ is the predicted label sequence for batch $j$, and $\mathcal{F}$ denotes the set of responses that follow the required format (e.g., tag correctness, answer presence, etc.).

    \item \textbf{Answer Uniqueness:}
    \[
    \text{Uniqueness} = \frac{1}{|\mathcal{D}|} \sum_{j=1}^{|\mathcal{D}|} \frac{|\text{set}(\hat{\mathbf{y}}^{(j)})|}{N_j}
    \]
    This metric quantifies how often the model avoids repeated predictions within a batch.
\end{itemize}

\subsection{Baselines}
\label{app_sec:baselines}

To contextualize the performance of \Ours, we compare it against a range of baseline models, covering both open- and closed-source LLMs. All baselines are evaluated using the same input format and candidate label set for fair comparison.

\textit{Zero-shot (with reasoning)}:
These models are prompted with our structured format, including metadata, gene expression, and candidate labels, but are not trained specifically on this task.

\begin{itemize}[leftmargin=*]
\item \texttt{LLaMA3.1-8B-Instruct}: A multilingual, instruction-tuned model with 8 billion parameters, optimized for dialogue tasks and outperforming many open and closed models on industry benchmarks.
\item \texttt{Qwen2.5-7B-Instruct}: A 7.61 billion parameter model, excelling in instruction following, long-text generation, and structured output, with enhanced capabilities in coding and mathematics.
\item \texttt{LLaMA3.3-70B-Instruct}: A 70 billion parameter instruction-tuned model, optimized for multilingual dialogue and outperforming many open and closed chat models on common industry benchmarks.
\item \texttt{GPT-4o}: OpenAI's flagship multimodal model capable of processing text, images, and audio, offering real-time responses and improved performance over its predecessors.
\item \texttt{GPT-4o-mini}: A smaller, more affordable version of \texttt{GPT-4o} by OpenAI, balancing performance and cost-efficiency for various AI applications.
\item \texttt{o3-mini}: OpenAI's cost-efficient reasoning model optimized for STEM tasks, particularly excelling in science, mathematics, and coding.
\item \texttt{o1}: OpenAI's model designed for complex reasoning and problem-solving tasks, employing a ``chain-of-thought'' approach to tackle multistep challenges effectively.
\end{itemize}

\textit{Instruction-tuned (without reasoning)}:
To test the value of reasoning, we fine-tune \texttt{LLaMA3.1-8B-Instruct} and \texttt{Qwen2.5-7B-Instruct} on the same training data but without reasoning traces. These models are directly optimized for answer-only classification.

All models are evaluated on the same held-out test set using five criteria: response length, cell-level accuracy, batch-level accuracy, output format correctness, and label uniqueness. The results are summarized in Table~\ref{tab:cell_annotation_results}.

\subsection{Implementation Details}
\label{app_sec:implem_details}

\paragraph{Supervised Fine-tuning.}
We fine-tune the model using the 3,912 distilled examples generated via reasoning distillation (see~\S\ref{sec:distill&sft}). Each instance is formatted using our standardized instruction prompt (see~\S\ref{sec:prompt_template}). We employ the \texttt{SFTTrainer} from the TRL library~\cite{vonwerra2022trl}, applying LoRA~\cite{hu2022lora} adaptation with rank $r=256$ to all linear layers. Training is conducted for 10 epochs with a learning rate of $5 \times 10^{-5}$, a batch size of 8, and gradient accumulation of 16 steps. The model backbone is \texttt{Qwen2.5-7B-Instruct}~\cite{qwen2.5, qwen2}, using left-side padding and truncation, and a maximum sequence length of 8,192 tokens. Evaluation is performed on 10\% of the training data at the end of each epoch.

\paragraph{Reinforcement Learning with GRPO.}
We combine the 3,912 fine-tuning instances with 3,000 additional instances to form an RL training set of 6,912 examples. Training is performed using the \texttt{verl}~\cite{sheng2024hybridflow} framework with GRPO~\cite{shao2024deepseekmath}. For each training input, 5 rollouts are sampled per prompt. The actor model is \texttt{Qwen2.5-7B-Instruct}, initialized from the SFT checkpoint.
We set the maximum prompt length to 3,072 tokens and response length to 3,000 tokens. The rollout engine is powered by \texttt{vLLM}~\cite{kwon2023efficient}, using 60\% GPU memory and tensor parallelism of 2. KL-regularization is applied using a low-variance formulation, with coefficient $\beta = 0.001$. The total batch size is 64 with micro batch size of 2 per GPU. Training is run for 20 epochs.

All experiments are conducted on 2 nodes with 8$\times$A100 GPUs (40GB).

\section{Cell-level vs. Batch-level Reasoning}

Our task formulation assumes joint reasoning over a batch of cells with shared context. To assess the necessity of this design, we introduce a decomposed variant in which each cell within a batch is annotated independently. Specifically, given a batch of $N$ cells and $N$ candidate labels, we split the task into $N$ individual classification instances—each involving one cell and the full candidate label set. While the label space remains the same, the model no longer has access to other cells in the batch during inference.

\begin{table}[h!]
    \centering
    \small
    \caption{Prompt template used for the decomposed cell-level reasoning task.}
    \label{tab:single_prompt_template}
    \vspace{2.5pt}
    \renewcommand{\arraystretch}{1.2}
    \begin{tabular}{p{13.5cm}}
        \hline
        You are an expert assistant specialized in cell type annotation. You will be given the gene expression profile of a single cell from a specific donor.
        The top expressed genes are listed in descending order. Use both gene expression and donor context to determine the correct cell type. \textbf{You will also receive a list of candidate cell types—choose the one that best fits this cell}. Include your detailed reasoning within \thinkbegin and \thinkend tags, and provide your final answer within \answerbegin and \answerend tags. The final answer should be a single string with exactly one cell type.
        \\
        \hline
    \end{tabular} 
\end{table}

We deliberately preserve the original candidate label set for each cell to retain the same classification difficulty, isolating the effect of batch-level context from label space complexity. 
In this setting, the prompt is constructed such that the model only receives a single cell's top expressed genes and metadata, along with a fixed set of candidate labels (see Table~\ref{tab:single_prompt_template}). Without access to other cells in the batch, the model cannot contrast expression patterns across candidates and must rely solely on local information to identify the correct label.

We consider two variants under this cell-level setup:
\textit{Cell-level Prediction Setup}: Instruction-tuned models that have not been exposed to any reasoning tasks. These models directly predict answers from the input without generating reasoning traces.
\textit{Cell-level Reasoning Setup}: Models capable of generating reasoning traces are evaluated on individual cells in isolation, requiring them to infer labels without the help of contextual comparisons among cells.

We evaluate both variants on 991 cells from 89 batches. As shown in Table~\ref{tab:single}, the cell-level prediction models ($^{\spadesuit}$) perform reasonably well in isolation (e.g., 47.1\% accuracy for Llama3.1), but fail to maintain consistency when reconstructed into batches (as low as 0\%\textasciitilde1.1\% batch-level accuracy). Cell-level reasoning models (e.g., \Ours) achieve higher cell-level accuracy (50.2\%), but still suffer a dramatic drop in batch-level accuracy (1.1\%) compared to their original joint reasoning setup. Even the strongest reasoning-capable models (e.g., OpenAI's \texttt{o1} and \texttt{o3-mini}) are unable to reliably solve the task when deprived of batch-level context, highlighting the inherent difficulty of single-cell reasoning in isolation.

\begin{table*}[h!]
\centering
\small
\footnotesize
\renewcommand\tabcolsep{4.5pt}
\renewcommand\arraystretch{1.05}
\caption{Comparison of model performance between batch-level and decomposed cell-level settings. $^{\spadesuit}$ indicates instruction-tuned baselines trained without any reasoning tasks, which directly predict answers from the input without generating reasoning traces. The best and second-best results are highlighted in \colorbox{backred!50}{red} and \colorbox{backblue!75}{blue}, respectively.}
\resizebox{0.93\textwidth}{!}{
\begin{tabular}{l|c|c|c|c|c}
\toprule
\header{Model} & \header{Response Length} & \header{Cell-level Acc} & \header{Batch-level Acc} & \header{Format} & \header{Uniqueness} \\
\midrule
\rowcolor[rgb]{0.93,0.93,0.93} \multicolumn{6}{c}{\textit{Cell-level Prediction Setup}} \\
\texttt{Llama3.1-8B-Instruct}$^{\spadesuit}$ & 8.1 & 0.4707 & 0.0112 & 0.6024 & 0.5887 \\
\texttt{Qwen2.5-7B-Instruct}$^{\spadesuit}$ & 8.5 & 0.3890 & 0.0000 & 0.5955 & 0.5978 \\
\rowcolor[rgb]{0.93,0.93,0.93} \multicolumn{6}{c}{\textit{Cell-level Reasoning Setup}} \\
\texttt{GPT-4o-mini} & 349.5 & 0.2761 &  0.0000 & \colorbox{backred!50}{1.0000} & 0.3706  \\
\texttt{GPT-4o} & 494.7 & 0.4093 & 0.0000 & \colorbox{backblue!75}{0.9900} & 0.5273 \\
\texttt{o3-mini} & 241.1 & 0.4825 & 0.0000 & \colorbox{backred!50}{1.0000} & 0.5663 \\
\texttt{o1} & 152.2 & 0.4700 & 0.0000 & \colorbox{backred!50}{1.0000} & 0.5708 \\
\Ours & 718.5 & 0.5016 &  0.0112 & 0.8427 & 0.6612 \\
\rowcolor[rgb]{0.93,0.93,0.93} \multicolumn{6}{c}{\textit{Batch-level Prediction Setup}} \\
\texttt{Llama3.1-8B-Instruct}$^{\spadesuit}$ & 73.8 & 0.6438 & 0.2809 & 0.9213 & 0.9198 \\
\texttt{Qwen2.5-7B-Instruct}$^{\spadesuit}$ & 78.7 & 0.5631 & 0.1910 & 0.9438 & 0.9370 \\
\rowcolor[rgb]{0.93,0.93,0.93} \multicolumn{6}{c}{\textit{Batch-level Reasoning Setup}} \\
\texttt{GPT-4o-mini} & 785.4 & 0.2330 & 0.0000 & \colorbox{backred!50}{1.0000} & 0.7147 \\
\texttt{GPT-4o} & 1010.4 & 0.4320 & 0.0000 & \colorbox{backred!50}{1.0000} &  0.8542 \\
\texttt{o3-mini} & 1059.9 & 0.6784 & 0.1461 & \colorbox{backred!50}{1.0000} & \colorbox{backred!50}{1.0000} \\
\texttt{o1} & 834.9 & \colorbox{backred!50}{0.7330} & \colorbox{backblue!75}{0.1685} & \colorbox{backred!50}{1.0000} & \colorbox{backblue!75}{0.9975} \\
\Ours & 1240.8 & \colorbox{backblue!75}{0.6953} & \colorbox{backred!50}{0.3371} & 0.9775 & 0.9759  \\
\bottomrule
\end{tabular}
}
\label{tab:single}
\end{table*}

In some cases, the model successfully identifies the correct labels in the batch-level reasoning setting, but fails in the cell-level reasoning setting when cells are presented individually.  
This discrepancy reveals that batch-level performance is not merely a function of memorizing cell types, but instead involves logical reasoning over inter-cell relationships. In the absence of such context, models are forced to “remember” what individual cell types look like based on internalized training knowledge—limiting their ability to disambiguate among fine-grained candidates.
These results highlight the critical role of structured context, which not only enforces consistency but also activates latent knowledge that may remain unused otherwise. The model is thus not merely retrieving memorized associations, but demonstrating an ability to perform inductive reasoning over a constrained hypothesis space when provided with contextual information.

\section{Open-ended QA vs. Constrained QA}

Although LLMs are naturally suited for open-ended question answering (QA), we find this formulation to be suboptimal for the task of cell type annotation. In the open-ended QA setup, the model is prompted to freely generate a cell type name for each cell in a given batch, based on its gene expression profile and metadata, without access to a constrained label set.

While this setting offers flexibility, it also introduces several challenges. Open-ended generation often results in inconsistent terminology or ambiguous outputs (e.g., ``T lymphocyte'' instead of ``CD4-positive, alpha-beta T cell''), making evaluation difficult and noisy. 
To mitigate this, we adopt the MedCPT~\cite{jin2023medcpt} score to quantify semantic similarity between the generated label and ground truth. For computing cell-level accuracy, we treat a prediction as correct if its MedCPT score exceeds 0.8. The MedCPT score reported in Table~\ref{tab:open} refers to the average similarity across all cells in the batch.

However, this threshold-based heuristic has notable limitations. It may conflate semantically similar yet biologically incorrect predictions with true positives, or penalize answers that are precise but phrased differently. 
As shown in Table~\ref{tab:open}, models like \texttt{o1} and \Ours achieve moderate MedCPT scores (e.g., 0.72\textasciitilde0.75), suggesting surface-level semantic alignment, but still exhibit 0\% batch-level accuracy. This gap underscores that semantic similarity does not imply coherent or consistent multi-cell reasoning.

In contrast, our constrained QA formulation provides the model with a predefined candidate label set, enabling structured reasoning and process-of-elimination strategies. This leads to significantly higher batch-level consistency and interpretability. For example, under constrained QA, \Ours achieves 32.9\% batch-level accuracy, far exceeding its open-ended counterpart.

Nevertheless, the constrained setup remains somewhat artificial, as the candidate label set is still narrower than what real-world annotation systems encounter. Future work should explore more realistic formulations by expanding the label set to better reflect the breadth of biomedical ontologies while retaining the benefits of constrained reasoning.

\begin{table*}[h!]
\centering
\small
\footnotesize
\renewcommand\tabcolsep{4.5pt}
\renewcommand\arraystretch{1.05}
\caption{Comparison of model performance between open-ended and constrained QA settings. $^{\spadesuit}$ indicates instruction-tuned baselines trained without any reasoning tasks, which directly predict answers from the input without generating reasoning traces. The best and second-best results are highlighted in \colorbox{backred!50}{red} and \colorbox{backblue!75}{blue}, respectively.}
\resizebox{0.93\textwidth}{!}{
\begin{tabular}{l|c|c|c|c|c}
\toprule
\header{Model} & \header{Response Length} & \header{Cell-level Acc} & \header{Batch-level Acc} & \header{Format} & \header{MedCPT Score} \\
\midrule
\rowcolor[rgb]{0.93,0.93,0.93} \multicolumn{6}{c}{\textit{Open-ended QA Setup}} \\
\texttt{Llama3.1-8B-Instruct}$^{\spadesuit}$ & 192.7 & 0.1068 & 0.0000 & 0.2822 & 0.7400 \\
\texttt{Qwen2.5-7B-Instruct}$^{\spadesuit}$ & 248.8 & 0.2491 & 0.0000 & 0.6804 & 0.7304 \\
\texttt{GPT-4o-mini} & 456.7 & 0.2326 & 0.0000 & 0.9973 & 0.6680 \\
\texttt{GPT-4o} & 401.0 & 0.3334 & 0.0000 & \colorbox{backblue!75}{0.9991} & 0.7060 \\
\texttt{o3-mini} & 590.7 & 0.4292 & 0.0000 & \colorbox{backred!50}{1.0000}  & 0.7411 \\
\texttt{o1} & 436.5 & 0.4411 & 0.0000 & \colorbox{backred!50}{1.0000} & 0.7519 \\
\Ours & 696.9 & 0.3367 & 0.0000 & 0.9224 & 0.7229 \\
\rowcolor[rgb]{0.93,0.93,0.93} \multicolumn{6}{c}{\textit{Constrained QA Setup}} \\
\texttt{Llama3.1-8B-Instruct}$^{\spadesuit}$ & 71.9 & 0.6411 & \colorbox{backblue!75}{0.2969} & 0.9078 & \colorbox{backblue!75}{0.9133} \\
\texttt{Qwen2.5-7B-Instruct}$^{\spadesuit}$ & 70.9 & 0.6179 & 0.2475 & 0.9443 & 0.8889 \\
\texttt{GPT-4o-mini} & 779.9 & 0.2581 & 0.0027 & 0.9918 & 0.7178 \\
\texttt{GPT-4o} & 1087.0 & 0.4248 & 0.0283 & 0.9817 & 0.7930 \\
\texttt{o3-mini} & 1130.2 & 0.5804 & 0.1352 & \colorbox{backred!50}{1.0000} & 0.8734 \\
\texttt{o1} & 906.6 & \colorbox{backblue!75}{0.6479} & 0.1900 & \colorbox{backred!50}{1.0000} & 0.8888 \\
\Ours & 1145.6 & \colorbox{backred!50}{0.6849} & \colorbox{backred!50}{0.3288} & 0.9826 & \colorbox{backred!50}{0.9185} \\
\bottomrule
\end{tabular}
}
\label{tab:open}
\end{table*}

\section{Human Evaluation Details}
\label{app_sec:evaluation}

We conducted a structured human evaluation to assess the reasoning quality of model predictions in the batch-level annotation task. The evaluation involved two domain experts with backgrounds in immunology and single-cell transcriptomics.

For each of the 100 randomly sampled test instances, annotators were shown model outputs including: (1) the predicted cell type for each cell in the batch, (2) the reasoning explanation provided for each prediction, and (3) a diagram indicating the annotation order and correctness of each step, as illustrated in Figure~\ref{fig:reasoning_analysis}. All outputs were anonymized and randomly shuffled to ensure a blind comparison between \Ours and baseline models.

Experts first determined whether each prediction was correct based on known marker expression and contextual alignment. For incorrect predictions, they further assigned one of the following predefined error types:

\begin{itemize}[leftmargin=*]

\item \textbf{Incorrect Reasoning}:  
  The explanation invokes wrong markers or faulty logic.\\
  \textit{Example: ``The cell highly expresses CD14 and LYZ, markers of T cells'', but CD14 and LYZ are actually monocyte/macrophage markers.}

\item \textbf{No Reasoning}:  
  A label is predicted with virtually no justification.\\
  \textit{Example: ``The cell is predicted to be a macrophage'', but nothing further.}

\item \textbf{Plausible but Incorrect}:  
  The reasoning sounds biologically sensible but ignores key evidence, resulting in an incorrect label.\\
  \textit{Example: ``The cell expresses CD3 and CD4, suggesting a helper T-cell identity'', but it also shows high levels of GZMB and NKG7—cytotoxic markers—so the true label is \textit{CD8+ cytotoxic T cell}.}

\item \textbf{Forced Assignment Error}:  
  The reasoning is correct, but to exhaust remaining labels, the model attaches the correct label to the wrong cell.\\
  \textit{Example: ``This is the only cell left that could be a monocyte'', even though an earlier cell was a better match.}

\item \textbf{Correct Reasoning, Wrong Conclusion}:  
  The explanation cites the right evidence, yet the model outputs a contradictory label.\\
  \textit{Example: ``The cell expresses CD3, CD4, and IL7R—consistent with helper T cells'', but the model outputs ``B cell''.}

\end{itemize}

All evaluations were performed independently, and ties were allowed when models performed comparably. The goal of the human study was to evaluate whether the model's reasoning steps reflected biologically grounded justifications and to identify common failure patterns in LLM predictions.

\section{Case Study: Comparing Reasoning Patterns Across Models}

To better understand model behaviors, we conduct qualitative comparisons of reasoning traces produced by different models on the same 10-cell batch (Tables~\ref{case:qwen}, \ref{case:4o}, \ref{case:o1}, and \ref{case:cello1}). We highlight correct predictions in \sethlcolor{lightblue}\hl{blue} and incorrect ones in \sethlcolor{lightred}\hl{red}, and analyze each model's reasoning style.

\subsection{\texttt{Qwen2.5-7B-Instruct}}
\texttt{Qwen2.5-7B-Instruct} (Table~\ref{case:qwen}) adopts a candidate-centric reasoning strategy, where it first enumerates all candidate cell types with brief gene-based descriptions and then loosely maps cells to the closest matching types. This approach relies on surface-level pattern matching between highly expressed genes and canonical cell type markers, such as matching TMSB4X to ``Intestine Goblet Cell'' and IGKC/JCHAIN to ``Plasma Cell.''

However, this model fails to leverage mutual exclusivity or batch-level constraints. For example, it assigns five different cells to ``Activated CD4-positive, Alpha-Beta T Cell,'' despite the task requiring one-to-one mapping between cells and types. Moreover, some predictions contradict the model's own reasoning: Cell 6 is labeled as ``Capillary Endothelial Cell'' even though it explicitly states that the evidence is ``not strongly supported.''

\begin{table}[h!]
    \centering
    \small
    \caption{Prediction of \texttt{Qwen2.5-7B-Instruct}.}
    \label{case:qwen}
    \vspace{2.5pt}
    \renewcommand{\arraystretch}{1.2}
    \begin{tabular}{p{13.5cm}}
    \toprule
    \textbf{Reasoning:} 
    Let's analyze the gene expression patterns and match them to the provided cell types. \\ 
    \textbullet\ Intestine Goblet Cell: This cell type is characterized by high expression of genes like TMSB4X, which is present in Cells 1, 3, 9. \\
    \textbullet\ Capillary Endothelial Cell: This cell type often expresses genes like CD34, CD31, and VWF. However, these genes are not listed here, so we cannot confidently assign this cell type based on the given data. \\
    \textbullet\ Plasma Cell: Plasma cells typically express genes like IGKC (Immunoglobulin kappa chain), JCHAIN, and IGHA1. These genes are present in Cell 8, suggesting it could be a plasma cell. \\
    \textbullet\ Memory B Cell: Memory B cells might express genes like CD74, but this is not prominently featured in any of the cells. \\
    \textbullet\ Activated CD4-positive, Alpha-Beta T Cell: This cell type would express genes like RPL41, RPL13A, and RPS27, which are common in Cells 2, 5, 6, 7, 10. \\
    \textbullet\ CD4-positive, Alpha-Beta T Cell: This cell type would also express genes like RPL41, RPL13A, and RPS27, similar to the activated CD4-positive, alpha-beta T cell. \\
    \textbullet\ T-helper 1 Cell: This cell type would express genes like IL2RA, IFNG, and STAT1, which are not present in the given gene lists. \\
    \textbullet\ Germinal Center B Cell: Germinal center B cells would express genes like CD74, CD21, and CD23, which are not prominently featured in any of the cells. \\
    \textbullet\ IgA Plasma Cell: This cell type would express genes like IGLL1, IGHG1, and IGHM, which are not present in the given gene lists. \\
    \textbullet\ Vein Endothelial Cell: This cell type would express genes like CD34, CD31, and VWF, which are not present in the given gene lists. \\
    \arrayrulecolor{gray!30}\midrule\arrayrulecolor{black}
    Based on the gene expression patterns, the most likely matches are: \\
    \textbullet\ \sethlcolor{lightred}\hl{Cell 1}: \sethlcolor{lightred}\hl{Intestine Goblet Cell} \\
    \textbullet\ \sethlcolor{lightred}\hl{Cell 2}: \sethlcolor{lightred}\hl{Activated CD4-positive, Alpha-Beta T Cell} \\
    \textbullet\ \sethlcolor{lightred}\hl{Cell 3}: \sethlcolor{lightred}\hl{Intestine Goblet Cell} \\
    \textbullet\ \sethlcolor{lightred}\hl{Cell 4}: \sethlcolor{lightred}\hl{Activated CD4-positive, Alpha-Beta T Cell} \\
    \textbullet\ \sethlcolor{lightred}\hl{Cell 5}: \sethlcolor{lightred}\hl{Activated CD4-positive, Alpha-Beta T Cell} \\
    \textbullet\ \sethlcolor{lightblue}\hl{Cell 6}: \sethlcolor{lightblue}\hl{Capillary Endothelial Cell (though not strongly supported)} \\
    \textbullet\ \sethlcolor{lightred}\hl{Cell 7}: \sethlcolor{lightred}\hl{Activated CD4-positive, Alpha-Beta T Cell} \\
    \textbullet\ \sethlcolor{lightred}\hl{Cell 8}: \sethlcolor{lightred}\hl{Plasma Cell} \\
    \textbullet\ \sethlcolor{lightred}\hl{Cell 9}: \sethlcolor{lightred}\hl{Intestine Goblet Cell} \\
    \textbullet\ \sethlcolor{lightred}\hl{Cell 10}: \sethlcolor{lightred}\hl{Activated CD4-positive, Alpha-Beta T Cell} \\
    \midrule
    \textbf{Answer:} \sethlcolor{lightred}\hl{intestine goblet cell} | \sethlcolor{lightred}\hl{activated cd4-positive, alpha-beta t cell} | \sethlcolor{lightred}\hl{intestine goblet cell} | \sethlcolor{lightred}\hl{activated cd4-positive, alpha-beta t cell} | \sethlcolor{lightred}\hl{activated cd4-positive, alpha-beta t cell} | \sethlcolor{lightblue}\hl{capillary endothelial cell} | \sethlcolor{lightred}\hl{activated cd4-positive, alpha-beta t cell} | \sethlcolor{lightred}\hl{plasma cell} | \sethlcolor{lightred}\hl{intestine goblet cell} | \sethlcolor{lightred}\hl{activated cd4-positive, alpha-beta t cell} \\
    \bottomrule
    \end{tabular} 
\end{table}
\begin{table}[h!]
    \centering
    \small
    \caption{Prediction of \texttt{GPT-4o}.}
    \label{case:4o}
    \vspace{2.5pt}
    \renewcommand{\arraystretch}{1.2}
    \begin{tabular}{p{13.5cm}}
    \toprule
    \textbf{Reasoning:} 
    To annotate the cell types, we need to analyze the gene expression profiles of each cell and match them to the most likely candidate cell type. The context of the donor (male, 13 years old, ileum, Crohn disease) is also relevant, as it may influence the types of cells present and their gene expression patterns. \\ 
    \arrayrulecolor{gray!30}\midrule\arrayrulecolor{black}
    \textit{Key Observations:} \\
    1. Gene Expression Patterns: \\
       \textbullet\ Many cells express ribosomal protein genes (e.g., RPLP1, RPL41, RPL10, RPS27, RPS18, etc.), which are common housekeeping genes. These genes alone are not sufficient for cell type identification. \\
       \textbullet\ Some cells express immune-related genes (e.g., IGKC, IGHA1, CD74, HLA-B, CXCR4), which suggest immune cell types. \\
       \textbullet\ Other cells express intestinal-specific genes (e.g., TFF3, SPINK4, REG1A, OLFM4), which are indicative of intestinal epithelial cells like goblet cells. \\
       \textbullet\ Endothelial-related genes (e.g., VWF, PLVAP, SPARC) are expressed in some cells, suggesting endothelial cell types. \\
    2. Candidate Cell Types: \\
       \textbullet\ The list includes immune cells (e.g., T cells, B cells, plasma cells), endothelial cells, and intestinal epithelial cells (e.g., goblet cells). \\
       \textbullet\ Some cell types are more specific (e.g., IgA plasma cell, activated CD4-positive T cell), while others are broader (e.g., plasma cell, vein endothelial cell). \\ 
    3. Contextual Relevance: \\
       \textbullet\ Crohn disease is an inflammatory bowel disease, so immune cells (e.g., T cells, B cells, plasma cells) and intestinal epithelial cells (e.g., goblet cells) are likely to be present. \\
       \textbullet\ The ileum is rich in immune cells (e.g., germinal center B cells, plasma cells) and specialized epithelial cells (e.g., goblet cells). \\
    \arrayrulecolor{gray!30}\midrule\arrayrulecolor{black}
    \textit{Cell-by-Cell Analysis:} \\
    \textbullet\ \sethlcolor{lightred}\hl{Cell 1}: Dominated by ribosomal genes and housekeeping genes (e.g., ACTB, MALAT1); No specific markers for immune or epithelial cells; Likely a \sethlcolor{lightred}\hl{generic plasma cell}. \\
    \textbullet\ \sethlcolor{lightred}\hl{Cell 2}: Similar to Cell 1, dominated by ribosomal genes; No strong markers for immune or epithelial cells; Likely a \sethlcolor{lightred}\hl{generic plasma cell}. \\
    \textbullet\ \sethlcolor{lightblue}\hl{Cell 3}: Expresses GAPDH and mitochondrial genes (e.g., MT-CO1, MT-CO2), along with ribosomal genes; No specific immune or epithelial markers; Likely a \sethlcolor{lightblue}\hl{generic plasma cell}. \\
    \textbullet\ \sethlcolor{lightblue}\hl{Cell 4}: Expresses intestinal-specific genes (e.g., TFF3, SPINK4, REG1A, OLFM4), which are markers for goblet cells; Likely an \sethlcolor{lightblue}\hl{intestine goblet cell}. \\
    \textbullet\ \sethlcolor{lightred}\hl{Cell 5}: Expresses mitochondrial genes (e.g., MT-CO1, MT-CO2) and ribosomal genes; No specific immune or epithelial markers; Likely a \sethlcolor{lightred}\hl{generic plasma cell}. \\
    \textbullet\ \sethlcolor{lightblue}\hl{Cell 6}: Expresses endothelial-related genes (e.g., VWF, PLVAP, SPARC) and immune-related genes (e.g., HLA-B, HLA-C); Likely a \sethlcolor{lightblue}\hl{capillary endothelial cell}. \\
    \textbullet\ \sethlcolor{lightblue}\hl{Cell 7}: Expresses endothelial-related genes (e.g., VWF, IGFBP7) and immune-related genes (e.g., HLA-B, HLA-E); Likely a \sethlcolor{lightblue}\hl{vein endothelial cell}. \\
    \textbullet\ \sethlcolor{lightblue}\hl{Cell 8}: Expresses immune-related genes (e.g., IGKC, IGHA1, JCHAIN), which are markers for IgA plasma cells; Likely an \sethlcolor{lightblue}\hl{IgA plasma cell}. \\
    \textbullet\ \sethlcolor{lightred}\hl{Cell 9}: Dominated by ribosomal genes and housekeeping genes; No specific immune or epithelial markers; Likely a \sethlcolor{lightred}\hl{generic plasma cell}. \\
    \textbullet\ \sethlcolor{lightblue}\hl{Cell 10}: Expresses immune-related genes (e.g., IGKC, IGHM, CD74, CXCR4), which are markers for memory B cells; Likely a \sethlcolor{lightblue}\hl{memory B cell}. \\
    \midrule
    \textbf{Answer:} \sethlcolor{lightred}\hl{Plasma cell} | \sethlcolor{lightred}\hl{Plasma cell} | \sethlcolor{lightblue}\hl{Plasma cell} | \sethlcolor{lightblue}\hl{Intestine goblet cell} | \sethlcolor{lightred}\hl{Plasma cell} | \sethlcolor{lightblue}\hl{Capillary endothelial cell} | \sethlcolor{lightblue}\hl{Vein endothelial cell} | \sethlcolor{lightblue}\hl{IgA plasma cell} | \sethlcolor{lightred}\hl{Plasma cell} | \sethlcolor{lightblue}\hl{Memory B cell} \\
    \bottomrule
    \end{tabular} 
\end{table}
\begin{table}[h!]
    \centering
    \small
    \caption{Prediction of \texttt{o1}.}
    \label{case:o1}
    \vspace{2.5pt}
    \renewcommand{\arraystretch}{1.2}
    \begin{tabular}{p{13.5cm}}
    \toprule
    \textbf{Reasoning:} 
    First, note that each of the 10 cells must map to a distinct one among the 10 candidate types. Some cells are straightforward to assign based on well-known marker genes: \\
    \textbullet\ \sethlcolor{lightblue}\hl{Cell 4} expresses TFF3, SPINK4, and REG1A, which are \sethlcolor{lightblue}\hl{classic intestinal goblet cell markers} (TFF3 is particularly specific). \\
    \textbullet\ \sethlcolor{lightblue}\hl{Cell 6} has PLVAP and VWF, both of which are strongly associated with endothelial cells—especially PLVAP, which is a hallmark of capillary (fenestrated) endothelium, so Cell 6 is a \sethlcolor{lightblue}\hl{capillary endothelial cell}. \\
    \textbullet\ \sethlcolor{lightblue}\hl{Cell 7} also expresses VWF but includes CCL21, HLA-E, and other features suggestive of a different endothelial subtype; since we must use both ``capillary endothelial cell'' and ``vein endothelial cell'', assigning Cell 7 to \sethlcolor{lightblue}\hl{vein endothelial cell} is consistent. \\
    \textbullet\ \sethlcolor{lightred}\hl{Cell 8} has IGKC, JCHAIN, and IGHA1 at high levels, unambiguously marking an IgA-expressing plasma cell \sethlcolor{lightred}\hl{(IgA plasma cell)}. \\
    \textbullet\ \sethlcolor{lightred}\hl{Cell 10} highly expresses IGKC and IGHM (and lacks IGHA1), indicating another antibody-producing cell but not IgA; this fits the generic \sethlcolor{lightred}\hl{``plasma cell''} category.
    \\
    \arrayrulecolor{gray!30}\midrule\arrayrulecolor{black}
    That leaves Cells 1, 2, 3, 5, and 9 to be mapped onto the remaining five types: CD4-positive, alpha-beta T cell; T-helper 1 cell; activated CD4-positive, alpha-beta T cell; memory B cell; germinal center B cell. \\
    \arrayrulecolor{gray!30}\midrule\arrayrulecolor{black}
    Although many of these cells' top 50 genes are dominated by ribosomal/housekeeping transcripts, we can still deduce plausible assignments. \sethlcolor{lightred}\hl{Cell 5}, for instance, is rich in mitochondrial genes (MT-CO1, MT-CO2, etc.) and also contains CD52 (commonly found on lymphocytes, especially T cells). High mitochondrial activity is characteristic of an activated T cell, so Cell 5 is best labeled \sethlcolor{lightred}\hl{``activated CD4-positive, alpha-beta T cell''}. 
    Cells 1, 3, and 9 are then the remaining T cells (two of them) plus one more B cell or vice versa. However, we need two B-cell categories overall—``memory B cell'' and ``germinal center B cell''—and we have not yet assigned either one to any cell besides the two plasma cells we already handled (Cells 8 and 10). Given that Cells 6 and 7 are endothelial, Cell 4 is goblet, Cell 8 and 10 are plasma lineage, and Cell 5 is activated T, we must place one B-cell type in Cell 2 and the other B-cell type in one of 1, 3, or 9.  \\
    \arrayrulecolor{gray!30}\midrule\arrayrulecolor{black}
    A close look at \sethlcolor{lightred}\hl{Cell 2} shows no strong expression of T-cell markers but instead a large suite of ribosomal transcripts, which can be consistent with a \sethlcolor{lightred}\hl{B lineage cell}. Meanwhile, among 1, 3, and 9, it is much more likely that two are T cells, and one is a B cell. The top genes in 1 and 3 both include TMSB4X at or near the top, which often appears in T cells (though it is not strictly T-specific, it is common in lymphocytes). Cell 9 is likewise heavy in housekeeping genes, and there is no direct B- or T-cell hallmark in the top transcripts. In practice, we must distribute the remaining categories such that we end up with exactly three T cells total and two non-plasma B cells total.  \\
    \arrayrulecolor{gray!30}\midrule\arrayrulecolor{black}
    A consistent assignment is:  \sethlcolor{lightred}\hl{Cell 1} $\rightarrow$ \sethlcolor{lightred}\hl{CD4-positive, alpha-beta T cell} (a ``baseline'' or less activated CD4 T cell); \sethlcolor{lightred}\hl{Cell 3} $\rightarrow$ \sethlcolor{lightred}\hl{T-helper 1 cell} (given the context of Crohn disease, Th1 skewing is common, and Cell 3 also has notable mitochondrial transcripts that can reflect an active state); Cell 5 (as reasoned above) $\rightarrow$ activated CD4-positive, alpha-beta T cell.  \\
    \arrayrulecolor{gray!30}\midrule\arrayrulecolor{black}
    That covers our three T-cell categories. For the two B-cell categories left (``memory B cell,'' ``germinal center B cell''), Cells 2 and 9 remain. Neither has overt immunoglobulin in the top 50, but we must use both B-cell categories somewhere. A typical assumption is that a germinal center B cell may show many ribosomal/housekeeping or proliferation genes, which could apply to Cell 9, while Cell 2 (also heavily ribosomal) could be the memory B cell. Ultimately, to satisfy the one-to-one requirement and given the absence of direct GC vs. memory markers at the top of the list, the most reasonable pairing is:  \sethlcolor{lightred}\hl{Cell 2} $\rightarrow$ \sethlcolor{lightred}\hl{memory B cell}; \sethlcolor{lightred}\hl{Cell 9} $\rightarrow$ \sethlcolor{lightred}\hl{germinal center B cell}.  \\ 
    \midrule
    \textbf{Answer:} \sethlcolor{lightred}\hl{CD4-positive, alpha-beta T cell} | \sethlcolor{lightred}\hl{memory B cell} | \sethlcolor{lightred}\hl{T-helper 1 cell} | \sethlcolor{lightblue}\hl{intestine goblet cell} | \sethlcolor{lightred}\hl{activated CD4-positive, alpha-beta T cell} | \sethlcolor{lightblue}\hl{capillary endothelial cell} | \sethlcolor{lightblue}\hl{vein endothelial cell} | \sethlcolor{lightblue}\hl{IgA plasma cell} | \sethlcolor{lightred}\hl{germinal center B cell} | \sethlcolor{lightred}\hl{plasma cell} \\
    \bottomrule
    \end{tabular} 
\end{table}
\begin{table}[h!]
    \centering
    \small
    \caption{Prediction of \Ours.}
    \label{case:cello1}
    \vspace{2.5pt}
    \renewcommand{\arraystretch}{1.2}
    \begin{tabular}{p{13.5cm}}
    \toprule
    \textbf{Reasoning:} 
    \textbullet\ \sethlcolor{lightblue}\hl{Cell 4} stands out immediately because of its strong expression of TFF3, SPINK4, REG1A, and PIGR—classic markers of intestinal goblet cells. Hence Cell 4 is the \sethlcolor{lightblue}\hl{``intestine goblet cell.''} \\
    \textbullet\ \sethlcolor{lightblue}\hl{Cell 8} has high levels of IGKC, JCHAIN, and IGHA1, which are hallmark genes of antibody-secreting plasma cells, especially IgA plasma cells. Thus Cell 8 is the \sethlcolor{lightblue}\hl{``IgA plasma cell.''} \\
    \textbullet\ \sethlcolor{lightblue}\hl{Cell 10} co-expresses IGKC and IGHM (among others), consistent with a B lineage cell that is not yet a fully differentiated plasma cell (since plasma cells typically downregulate IGHM). This fits best with a ``memory B cell'' or a ``germinal center B cell.'' We will see that Cell 10 is most consistent with a \sethlcolor{lightblue}\hl{``memory B cell''} once we compare it with Cell 8 and Cell 1. \\
    \textbullet\ \sethlcolor{lightblue}\hl{Cell 6} has prominent expression of PLVAP, VWF, and SPARC. PLVAP is a well-known marker of fenestrated capillary endothelium, and VWF is a canonical endothelial marker. This makes Cell 6 the \sethlcolor{lightblue}\hl{``capillary endothelial cell.''}
    \\
    \textbullet\ \sethlcolor{lightblue}\hl{Cell 7} also shows strong expression of VWF and CCL21. CCL21 is famously associated with lymphatic endothelium, and VWF is again a hallmark of endothelium. This places Cell 7 as the \sethlcolor{lightblue}\hl{``vein endothelial cell''} (rather than capillary), leaving Cell 6 to be the capillary endothelial cell. \\
    \textbullet\ Cell 1 and Cell 2 remain unassigned. Both are dominated by ribosomal genes and do not show overt markers for T cells, B cells, or endothelial cells. However, we must use the remaining two cell types: ``CD4-positive, alpha-beta T cell'' and ``activated CD4-positive, alpha-beta T cell.'' Neither Cell 1 nor Cell 2 shows strong evidence of activation (e.g. no PFN1 in Cell 2), so \sethlcolor{lightred}\hl{Cell 1} is most likely the \sethlcolor{lightred}\hl{``activated CD4-positive, alpha-beta T cell,''} and \sethlcolor{lightblue}\hl{Cell 2} is the \sethlcolor{lightblue}\hl{``CD4-positive, alpha-beta T cell.''} \\
    \textbullet\ \sethlcolor{lightred}\hl{Cell 3} is the remaining unassigned cell. The only remaining type is ``germinal center B cell,'' so Cell 3 must be the \sethlcolor{lightred}\hl{``germinal center B cell.''} \\
    \arrayrulecolor{gray!30}\midrule\arrayrulecolor{black}
    Putting these all together gives: \\
    \textbullet\ \sethlcolor{lightred}\hl{Cell 1} $\rightarrow$ \sethlcolor{lightred}\hl{activated CD4-positive, alpha-beta T cell}  \\
    \textbullet\ \sethlcolor{lightblue}\hl{Cell 2} $\rightarrow$ \sethlcolor{lightblue}\hl{CD4-positive, alpha-beta T cell} \\ 
    \textbullet\ \sethlcolor{lightred}\hl{Cell 3} $\rightarrow$ \sethlcolor{lightred}\hl{germinal center B cell}  \\
    \textbullet\ \sethlcolor{lightblue}\hl{Cell 4} $\rightarrow$ \sethlcolor{lightblue}\hl{intestine goblet cell}  \\
    \textbullet\ \sethlcolor{lightred}\hl{Cell 5} $\rightarrow$ \sethlcolor{lightred}\hl{CD4-positive, alpha-beta T cell with a helper phenotype} (i.e. ``T-helper 1 cell,'' even though the T-helper 1 signature genes are not in the top 50).  \\
    \textbullet\ \sethlcolor{lightblue}\hl{Cell 6} $\rightarrow$ \sethlcolor{lightblue}\hl{capillary endothelial cell}  \\
    \textbullet\ \sethlcolor{lightblue}\hl{Cell 7} $\rightarrow$ \sethlcolor{lightblue}\hl{vein endothelial cell}  \\
    \textbullet\ \sethlcolor{lightblue}\hl{Cell 8} $\rightarrow$ \sethlcolor{lightblue}\hl{IgA plasma cell}  \\
    \textbullet\ \sethlcolor{lightred}\hl{Cell 9} $\rightarrow$ a less activated or naive/naïve-like T/B lineage cell, but we already used the other T-cell categories.  \\
    \textbullet\ \sethlcolor{lightblue}\hl{Cell 10} $\rightarrow$ \sethlcolor{lightblue}\hl{memory B cell}  \\
    \arrayrulecolor{gray!30}\midrule\arrayrulecolor{black}
    However, we notice that Cell 5 does not fit the ``activated'' or ``memory'' categories. It does have CD52 (commonly found on T cells) and PFN1 (also can be a marker of helper T cells). Since we must place it as a helper T cell, Cell 5 is the ``T-helper 1 cell.''
    Hence the final mapping is: \\
    \textbullet\ \sethlcolor{lightred}\hl{Cell 1} $\rightarrow$ \sethlcolor{lightred}\hl{activated CD4-positive, alpha-beta T cell}  \\
    \textbullet\ \sethlcolor{lightblue}\hl{Cell 2} $\rightarrow$ \sethlcolor{lightblue}\hl{CD4-positive, alpha-beta T cell} \\ 
    \textbullet\ \sethlcolor{lightred}\hl{Cell 3} $\rightarrow$ \sethlcolor{lightred}\hl{germinal center B cell}  \\
    \textbullet\ \sethlcolor{lightblue}\hl{Cell 4} $\rightarrow$ \sethlcolor{lightblue}\hl{intestine goblet cell}  \\
    \textbullet\ \sethlcolor{lightblue}\hl{Cell 5} $\rightarrow$ \sethlcolor{lightblue}\hl{T-helper 1 cell}   \\
    \textbullet\ \sethlcolor{lightblue}\hl{Cell 6} $\rightarrow$ \sethlcolor{lightblue}\hl{capillary endothelial cell}  \\
    \textbullet\ \sethlcolor{lightblue}\hl{Cell 7} $\rightarrow$ \sethlcolor{lightblue}\hl{vein endothelial cell}  \\
    \textbullet\ \sethlcolor{lightblue}\hl{Cell 8} $\rightarrow$ \sethlcolor{lightblue}\hl{IgA plasma cell}  \\
    \textbullet\ \sethlcolor{lightred}\hl{Cell 9} $\rightarrow$ \sethlcolor{lightred}\hl{activated CD4-positive, alpha-beta T cell (but we only have one ``activated'' category), so we assign it to the remaining ``plasma cell'' category.} \\
    \textbullet\ \sethlcolor{lightblue}\hl{Cell 10} $\rightarrow$ \sethlcolor{lightblue}\hl{memory B cell}  \\
    \midrule
    \textbf{Answer:} 
    \sethlcolor{lightred}\hl{activated CD4-positive, alpha-beta T cell} | \sethlcolor{lightblue}\hl{CD4-positive, alpha-beta T cell} | \sethlcolor{lightred}\hl{germinal center B cell} | \sethlcolor{lightblue}\hl{intestine goblet cell} | \sethlcolor{lightblue}\hl{T-helper 1 cell} | \sethlcolor{lightblue}\hl{capillary endothelial cell} | \sethlcolor{lightblue}\hl{vein endothelial cell} | \sethlcolor{lightblue}\hl{IgA plasma cell} | \sethlcolor{lightred}\hl{plasma cell} | \sethlcolor{lightblue}\hl{memory B cell} \\
    \bottomrule
    \end{tabular} 
\end{table}

\subsection{\texttt{GPT-4o}}
\texttt{GPT-4o} (Table~\ref{case:4o}) displays strong domain knowledge and an ability to reason using both gene expression patterns and contextual biological information. It explicitly incorporates factors such as donor condition (Crohn disease), anatomical site (ileum), and immune cell enrichment. It also organizes its reasoning with structured sections: \textit{general observations}, \textit{candidate type definitions}, and \textit{cell-by-cell analysis}, showing an impressive degree of language-level coherence.

Despite this, GPT-4o suffers from constraint-agnostic generation. It frequently assigns the same label (e.g., ``Plasma cell'') to multiple cells—violating the task’s one-to-one mapping requirement. This behavior suggests the model treats the task as a sequence of independent classification problems rather than a structured inference task over a constrained hypothesis space. 

Furthermore, although its biological interpretation is often correct (e.g., identifying Cell 4 as a goblet cell, Cell 8 as an IgA plasma cell), the model overuses broad fallback categories like ``generic plasma cell'' in ambiguous cases, indicating uncertainty resolution by generalization rather than reasoning.

\subsection{\texttt{o1}}
The \texttt{o1} model exhibits a clear understanding of the one-to-one mapping constraint inherent to the task. It explicitly structures its reasoning in two stages: (1) assigning obvious cells based on canonical markers, and (2) carefully distributing the remaining ambiguous cells to satisfy label exclusivity. For example, Cell 4 is confidently identified as an ``Intestine Goblet Cell'' based on strong markers like TFF3 and REG1A, while Cells 6 and 7 are assigned to endothelial subtypes based on PLVAP and VWF expression.

What sets \texttt{o1} apart is its deliberate reasoning over cell allocation: it acknowledges that only one cell can be assigned to each of the B-cell and T-cell subtypes, and explicitly discusses the combinatorial implications (e.g., ``we must distribute the remaining categories such that we end up with exactly three T cells and two non-plasma B cells''). This illustrates a form of global, constraint-aware reasoning that is absent in instruction-tuned or open-ended models.

However, while the logical structure is robust, the model still makes several incorrect assignments. For example, Cell 1 is predicted as a ``CD4-positive, alpha-beta T cell,'' though it likely belongs to another lineage; Cell 2 is assigned to ``memory B cell'' based on ribosomal gene dominance rather than immunoglobulin expression. These errors reflect plausible but weakly supported conclusions, often stemming from ambiguity in gene expression or over-reliance on inferred distributions rather than molecular markers.

\subsection{\Ours}

\Ours demonstrates the strongest reasoning ability among all models in the case study. Its inference is explicitly structured in a progressive assignment strategy, beginning with high-confidence matches based on canonical markers (e.g., Cell 4 as ``intestine goblet cell'' via TFF3/REG1A; Cell 8 as ``IgA plasma cell'' via IGKC/JCHAIN/IGHA1) before moving on to more ambiguous cases.

A key strength of \Ours lies in its explicit modeling of mutual exclusivity and type constraints. The model carefully tracks which labels remain unassigned and distributes ambiguous cells accordingly. For example, it defers the assignment of Cell 1 and Cell 2 until later in the reasoning chain, acknowledging the overlap in ribosomal profiles and the lack of activation markers, before settling on the most consistent label-to-cell pairing.

Importantly, \Ours also performs cross-cell comparisons (e.g., comparing Cell 10 to Cell 8 to support a ``memory B cell'' label), showing a depth of relational reasoning absent from previous models. It further revisits earlier assignments (e.g., Cell 5), updating the label from a generic CD4-positive T cell to a ``T-helper 1 cell'' based on expression of CD52 and PFN1.

However, the model still makes minor errors. Notably, Cell 3 is assigned to ``germinal center B cell'' with minimal molecular justification, and Cell 9 is incorrectly labeled as ``plasma cell'' due to the unavailability of any better-fitting categories—this is acknowledged in the reasoning as a forced fallback. These mistakes are not due to poor logic but rather task constraints interacting with biological ambiguity.

\end{document}